\documentclass[bookmarks=true, hyperfootnotes=false]{article}
\usepackage{ijcai26}

\usepackage{times}
\usepackage{soul}
\usepackage{url}
\usepackage[hidelinks]{hyperref}
\usepackage[utf8]{inputenc}
\usepackage[small]{caption}
\usepackage{graphicx}
\usepackage{amsmath}
\usepackage{amsthm}
\usepackage{booktabs}
\usepackage{algorithm}
\usepackage{algorithmic}
\usepackage[switch]{lineno}
\usepackage{array}

\usepackage[utf8]{inputenc} 
\usepackage[T1]{fontenc}    
\usepackage{hyperref}       
\usepackage{url}            
\usepackage{booktabs}       
\usepackage{amsfonts}       
\usepackage{nicefrac}       
\usepackage{microtype}      
\usepackage{xcolor}         

\usepackage{natbib}
\setcitestyle{authoryear,open={},close={]},citesep={;}}
\usepackage{multicol}
\usepackage{graphicx}
\usepackage[symbol]{footmisc}
\usepackage{tablefootnote}
\usepackage{caption}
\usepackage{subcaption}
\usepackage{todonotes}

\usepackage{xcolor}

\DeclareRobustCommand{\legendsquare}[1]{%
  \textcolor{#1}{\rule{1ex}{1ex}}%
}
\definecolor{strongDisColor}{HTML}{FFB000}
\definecolor{disColor}{HTML}{FE6100}
\definecolor{neuColor}{HTML}{DC267F}
\definecolor{agreeColor}{HTML}{785EF0}
\definecolor{strongAgreeColor}{HTML}{648FFF}

\makeatletter
\newcommand\footnoteref[1]{\protected@xdef\@thefnmark{\ref{#1}}\@footnotemark}
\makeatother

\begin{document}
\title{How Users Understand Robot Foundation Model Performance through Task Success Rates and Beyond}

\author{
Isaac Sheidlower$^1$\and
Jindan Huang$^2$\and
James Staley $^2$\and
Bingyu Wu $^2$\and
Qicong Chen $^2$\and \\
Reuben M Aronson $^2$\And
Elaine Short $^2$\\
\affiliations
$^1$Brown University, Providence, RI, USA\\
$^2$Tufts University, Medford, MA, USA\\
\emails{
Isaac\_Sheidlower@Brown.edu, jindan.huang@tufts.edu, james.staley625703@tufts.edu, bingyu.wu@tufts.edu, qicong.chen@tufts.edu, reuben.aronson@gmail.com, elaine.short@tufts.edu
}
}
\maketitle
\begin{abstract}
   Robot Foundation Models (RFMs) represent a promising approach to developing general-purpose home robots. Given the broad capabilities of RFMs, users will inevitably ask an RFM-based robot to perform tasks that the RFM was not trained or evaluated on. In these cases, it is crucial that users understand the risks associated with attempting novel tasks due to the relatively high cost of failure. Furthermore, an informed user who understands an RFM's capabilities will know what situations and tasks the robot can handle. In this paper, we study how non-roboticists interpret performance information from RFM evaluations. These evaluations typically report task success rate (TSR) as the primary performance metric. While TSR is intuitive to experts, it is necessary to validate whether novices also use this information as intended. Toward this end, we conducted a study in which users saw real evaluation data, including TSR, failure case descriptions, and videos from multiple published RFM research projects. The results highlight that non-experts not only use TSR in a manner consistent with expert expectations but also highly value other information types, such as failure cases that are not often reported in RFM evaluations. Furthermore, we find that users want access to both real data from previous evaluations of the RFM and estimates from the robot about how well it will do on a novel task.
\end{abstract}

\section{Introduction}
\begin{figure}[t]
\centering
\includegraphics[width=.48\textwidth]{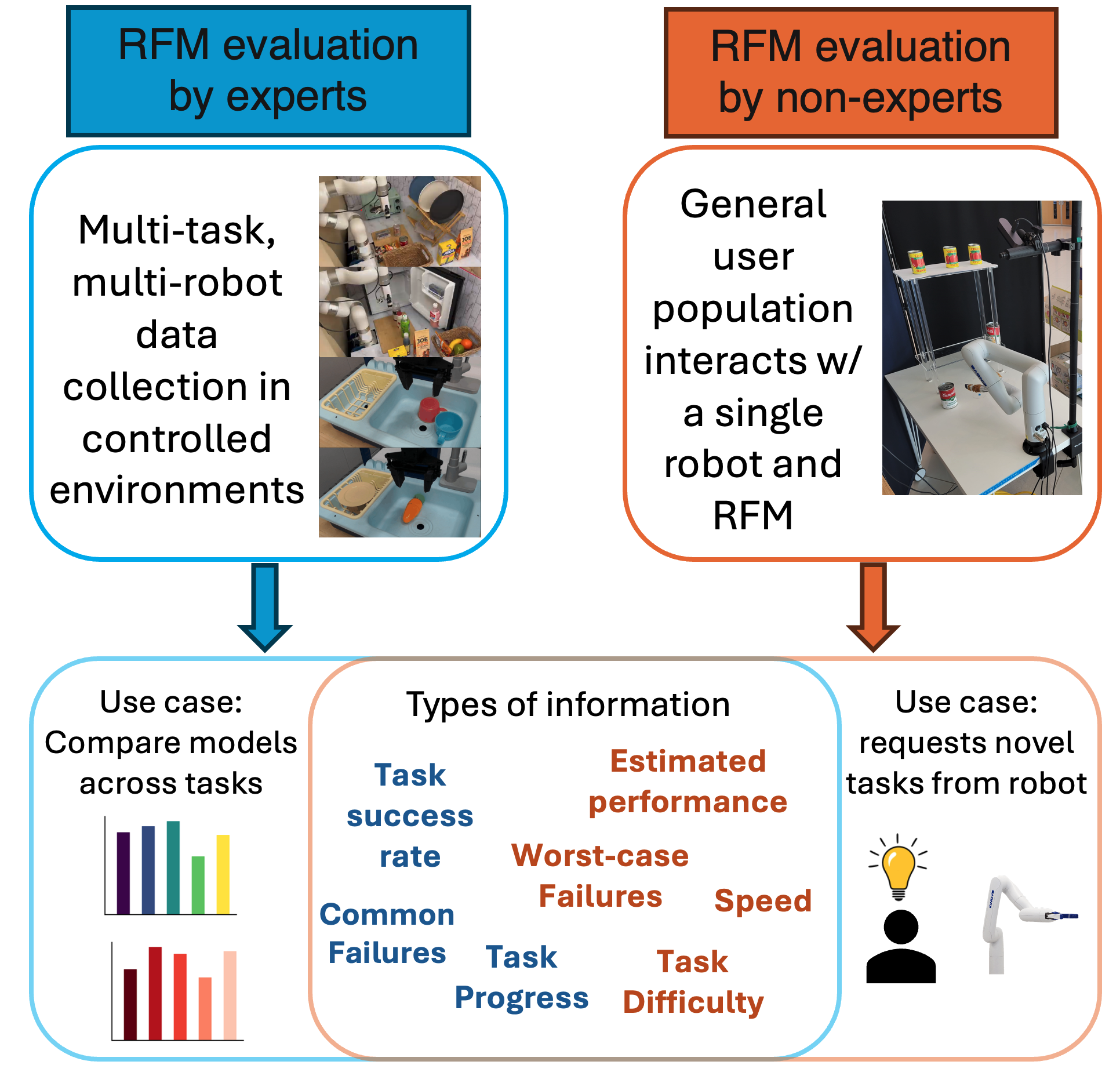}

\vspace*{-3mm}

\caption{This work investigates how people interpret commonly reported robot foundation model performance information. We verify that non-experts value information like task success rate, while also wanting other information types less readily available. This work serves to inform future robot foundation model evaluations as well as what information should be available to end-users during deployments and when they request tasks.}%
\vspace{-6mm}
\label{fig:firstPage}
\end{figure}

Robot Foundation Models (RFMs) promise that users can request robots to perform nearly arbitrary tasks to an acceptable standard. To realize this promise safely and effectively, users must be informed about RFM performance on such tasks, both to avoid risky or catastrophic failures and to understand when, how, and for which tasks a robot can be used successfully. The most prevalent evaluation metric for conveying this performance is \textit{task success rate} (TSR), defined as the ratio of successful task executions to total attempts. TSR is often interpreted as an indication of how likely the robot is to succeed at that task. TSR is useful for novices and experts alike: experts can use TSR to compare the performance of an RFM on different tasks, while novices may find TSR particularly useful when requesting an RFM-based robot to perform a novel task, or a task that the user has not seen the robot perform. If the robot provides an accurate estimate of the TSR, or a TSR based on a very similar task, then the user can better decide how to use the robot. 

Consider the task of putting a vase away on a shelf: if the TSR on that task is very high, the person can confidently request the robot to perform that task with little supervision. If the TSR is very low, the person may choose to closely monitor the robot, further teach the robot, or decide it should not attempt the task at all. While TSR is the most commonly-used metric for assessing RFMs, other information about robot performance can be helpful. \textit{Failure cases}, which describe how a robot failed in natural language, are less commonly reported in RFM evaluations but are useful for predicting and preparing for failures. For example, if the robot explains that dropping the vase is a possible failure case, then the user may closely monitor the robot and be prepared to catch the vase. If the user is not provided with this information, then the robot may fail unexpectedly, causing both a broken vase and potentially broken trust in the robot in general. The more informed a user is about an RFM's capabilities, the better they can use it for routine and creative tasks alike. 

In this work, we study how \textbf{novice} users interact with RFM evaluations from real robots, focusing on their interpretations of TSR and failure cases. Our goal is to understand what information users need when using RFM-based robots for novel tasks, how they interpret commonly reported evaluation data, and where gaps exist between reported and user-desired information. We focus on scenarios in which users request tasks they have not seen the robot execute and that the RFM has not previously attempted, requiring information to compensate for this lack of experience. We conducted an online study (\textbf{n}=112) using real RFM evaluation data, and an in-person study (\textbf{n}=14) with an autonomous robot to examine how embodiment affects information needs.

Our contributions are: \textbf{(1)} We empirically validate that non-expert users interpret TSR in a manner consistent with expert expectations, demonstrating a robust positive correlation between TSR values and user-reported trust and confidence in the system. \textbf{(2)} We show that natural language descriptions of failure cases are highly valued by non-expert users because they further inform their understanding of the risks associated with using the robot, suggesting a need to standardize failure case reporting. We also provide recommendations for information types that should be available during RFM deployments based user data, such as the speed at which a task can be completed and the RFM's propensity for human-robot collaborations. \textbf{(3)} Finally, we highlight the necessity of providing users both with \emph{real} evaluation data of similar tasks to the ones users request, suggesting the need for RFMs to be deployed with access to a large history of evaluations, and provide a means for the RFM to accurately \emph{estimate} its performance on a task it has not yet been evaluated for. In other words, RFMs should be intentionally designed to behave predictably on new tasks, either by exhibiting similar performance to that demonstrated in previous related tasks or by giving users a way to visualize or forecast their behavior. Together, these findings motivate a shift toward user-centered, interpretable evaluations and deployments of RFM based robots. 

\section{Related Works}
In this work, an RFM refers to a general-purpose model or algorithm which can be deployed on a robot enabling it to perform a wide range of physical manipulation tasks as specified by a user. RFMs often manifest as ``generalist policies'' or ``vision language action models (VLAs)'', and are gaining popularity as a way to distill large amounts of data from a wide range of tasks into a single model; RFMs, such as RT-2 [\citep{brohan_rt-2_2023} and OpenVLA [\citep{kim_openvla_2024}, are  end-to-end policies that take language instructions or goal images along with camera observations as input, and output actions, usually in end-effector space.  Like LLM model scaling [\citep{zhao_survey_2024}, more diverse data and larger models tend to improve an RFMs performance and generalization capabilities.  The success of RFMs has also coincided with the advancement of data-efficient and lower cost imitation learning methods [\citep{chi_diffusion_2024, zhao_learning_2023, team_aloha_2024} and smaller-scale but effective multi-task learning techniques [\citep{shridhar_perceiver-actor_2023, wang_sparse_2024, reuss_multimodal_2024, haldar_baku_2024}. Given that these models are still in their early days, there is still relatively little work on how to best holistically evaluate them.

RFMs are typically evaluated with TSR across a variety of different tasks: typically short-horizon pick-and-place tasks conducted with robot arms. Although many RFMs are evaluated on real robots, recent advances have also enabled high-fidelity simulation evaluations [\citep{liu_libero_2024}. TSR is useful for enabling direct comparisons of different RFMs on the same tasks; however, it is not the only means of evaluating these models and is often overemphasized in evaluation. For example, some works also discuss \emph{failure cases} for tasks. Despite the impressive capabilities of RFMs, they are still very prone to failing; though less commonly reported, how they fail can be useful information for researchers [\citep{liu_reflect_2023, duan_aha_2024} and potentially novice users alike. Recent work by Wang et al. [\citeyearpar{wang_ladev_2024, wang_towards_2024} proposes benchmarks for evaluating RFMs in simulation wherein tasks are broken down into sub-tasks and TSR is reported for each (e.g. a pick-and-place task requires the robot to move to the correct object and successfully grasp it before moving it to a different location). Estimating TSR and if/how a robot will fail is also an active area of research [\citep{goko_task_2024, diehl_causal-based_2023}. Despite this progress, the performance of RFMs has yet to be evaluated and studied with novices, to the best of our knowledge. 

In the field of human-robot interaction (HRI), it is well studied and understood that users who are informed about a robot and its capabilities improve their ability to collaborate with, teach, and appropriately trust a robot [\citep{stubbs_autonomy_2007, pandya_multi-agent_2024, iucci_explainable_2021, moorman_investigating_2023}. Failure can significantly impact a user's trust in a robot and how it impacts that person can vary depending on both the nature of the task and the user [\citep{khavas_review_2021}. Specific types of robot failures and their severity can alter how a user interacts with a robot, for example by influencing their chosen method of teaching it [\citep{huang_effect_errors_2024}. TSR, on the other hand, can be interpreted as reflecting the probability of task success. People's interpretation of probabilities can vary greatly depending on how those probabilities are presented [\citep{oudhoff_effect_2015, chater_probabilistic_2008}. Attending to communication of risk and probabilities is well-studied in the high-stakes healthcare domain, where it is critical to ensure patients can make informed decisions and has consequently been widely studied  [\citep{ancker_scope_2025, johnson_presenting_1995}; these same principles apply to the high-stakes robotics domain, wherein people share physical space with robots. Thus, it is crucial that we investigate the gap between how non-experts understand and the way experts understand and use TSR.

\section{Methodology: Defining information types} 
To identify how non-experts interpret and use performance information, we designed an online study where participants were presented with real RFM evaluation data, asked to evaluate a robot's capabilities and to brainstorm what information types would be useful when using such a robot. We then conducted an in-person study to corroborate these findings with embodiment. Here we describe the four information types we studied. These types are drawn from the commonly reported ``task success rate'' and ``failure case'' information types found in RFM evaluations. Each is presented with an example from the study using the task request ``Move the green bottle of tea to the refrigerator door'' as reference. 

\textbf{Estimated Task Success Rate (ETSR)}
This refers to a robot's internal estimate of how likely it is to perform a requested task successfully. This functionality is likely crucial for robots to be deployed safely. Estimating a robot's ability to perform a given task is an ongoing area of research. \textit{E.g.: ``The robot estimates it can successfully complete the task 1/5 times, or 20\% of the time.''}

\textbf{Estimated Failure Case (EFC)}
This refers to a robot's estimate of how it may fail when performing a task. Similar to ETSR, this type of functionality has both safety and usability implications. We focus on EFC as a natural language description (vs.~a failure example shown in a simulator). Furthermore, we treat this estimate as the most likely failure case. This may not be the only useful failure case; a ``worst case'' failure scenario, for example, may also be important to users. We choose to present EFC to participants in the framing as the ``most likely failure case,'' as such cases are often discussed in research. \textit{E.g.: ``The robot estimates that it may fail by attempting to pick up an object that is not the green tea.''}

\textbf{Related Task - Task Success Rate (RT-TSR)}
This refers to previously collected task data from an RFM and its resulting TSR. RT-TSR is TSR data from a task that is ``similar'' to the user's requested task. Similarity can mean different things and be measured in different ways (see Section III, B for how we measure it in this study). Unlike ETSR, RT-TSR is derived from real roll-outs of the RFM and users may value it differently than a robot's internal estimate. RT-TSR can also be derived from RFM deployments in different robots and environments. Although the performance of an RFM on one task may not always be indicative of how it performs on another, RT-TSR provides a grounded data point for understanding an RFM's capabilities. \textit{E.g.: ``The robot has previously succeeded in this similar task 4/5 times, or 80\% of the time.''}

\textbf{Related Task - Failure Case (RT-FC)}
This refers to real failure cases that occurred in similar robot tasks. Like EFC, we focus on the case of a verbal description of the most likely failure for that task. RT-FC can be useful for gauging how an RFM may be prone fail. For example, in a pick-and-place task, an RFM failing to identify the right object to pick may indicate something different than an RFM failing to grasp the requested object. \textit{E.g.: ``The robot has previously failed this similar task by grasping the can but then halting, failing to put it in the refrigerator.''}

\begin{center}
\begin{table}[t]{%
\scriptsize
\begin{tabular}{|l|l|l|}
\hline
\textbf{Model} & \textbf{\begin{tabular}[c]{@{}l@{}}Robots Used\end{tabular}} & \textbf{Tasks}                     \\ \hline
\begin{tabular}{@{}l@{}}
    OpenVLA \\ \phantom
      [ [\citep{kim_openvla_2024} \tablefootnote[1]{\label{note1} We thank the authors of this work for generously providing us with the evaluation data and videos upon our request.}\
\end{tabular}&
  \begin{tabular}[c]{@{}l@{}}WidowX, \\ Franka Emika Panda\end{tabular} &
  \begin{tabular}[c]{@{}l@{}}Remove battery from sink;\\ Move salt shaker onto plate;\\ Move eggplant from sink to pot;\\ Move carrot from sink to plate;\\Stack cups\end{tabular} \\ \hline
\begin{tabular}{@{}l@{}}
    Baku \\ \phantom  [ [\citep{haldar_baku_2024} 
\end{tabular} &
  Ufactory xArm &
  \begin{tabular}[c]{@{}l@{}}Remove can from refrigerator; \\ Close oven door; Lift lid off pan; \\ Lift orange out of bowl;\\ Put coke can in basket; \\ Wipe cutting board;\\ Move tea bottle to refrigerator\end{tabular} \\ \hline
\begin{tabular}[c]{@{}l@{}}Multimodal \\ Diffusion\\ Transformer (MDT) \\ \phantom [[ \citep{reuss_multimodal_2024}\footnotemark{} \end{tabular} &
  Franka Emika Panda &
  \begin{tabular}[c]{@{}l@{}}Move banana from stove to sink;\\ Move banana from sink to stove;\\ Move pot from sink to stove;\\ Push toaster lever down\end{tabular} \\ \hline
\end{tabular}%
}
\vspace{-3mm}
\caption{Models, robots, and tasks shown to users.}
\vspace{-5mm}
\label{table:models}
\end{table}
\vspace{-5mm}
\end{center}

\section{User Experience with Different Information Types (online study)}
\subsection{Procedure}
\textbf{Task Data Collection and Coding}
We collected real RFM data through surveying RFM literature and through collaborations with authors of RFM and multi-task robot learning works. We collected task data that contained: TSR as collected from an RFM evaluation, including the number of total trials; a video of the robot successfully performing a task; and a video of the robot failing a task if the task success rate was below $100\%.$ This resulted in 16 tasks from three evaluations (see Table 1). All participant-facing material used in the study was from robot evaluations in published works. The data used for ETSR and EFC \emph{was real data}, but was framed as estimates to participants. This ensured that ETSR and EFC were accurate and could be presented without developing an estimator. The mean TSR was approximately 66\% (SD $\approx$ 23\%).

The study required pairs of similar tasks to present RT-TSR and RT-FC. We tested several methods for encoding task similarity, such as the distance of task descriptions in language embedding space and querying LLMs. Of the methods, qualitative coding [\citep{auerbach_qualitative_2003} was the most consistent. We had three members of our lab, each of whom was unaware of the study's content, independently label the most similar task for all 16 tasks. Similarity was based on the ``robot skills'' required to complete each task given a task description and the failure and success videos. Disagreements were resolved through discussion. We wrote descriptions of the failure cases in the videos, which were independently checked by two researchers for accuracy and objectivity. We intentionally avoided any speculation as to the cause of the failure, but rather attempted to just describe what physically took place. See the appendix for the failure descriptions, task similarity coding results, and a complete list of all TSRs.

\begin{figure*}[t]
\centering
\includegraphics[width=.94\textwidth]{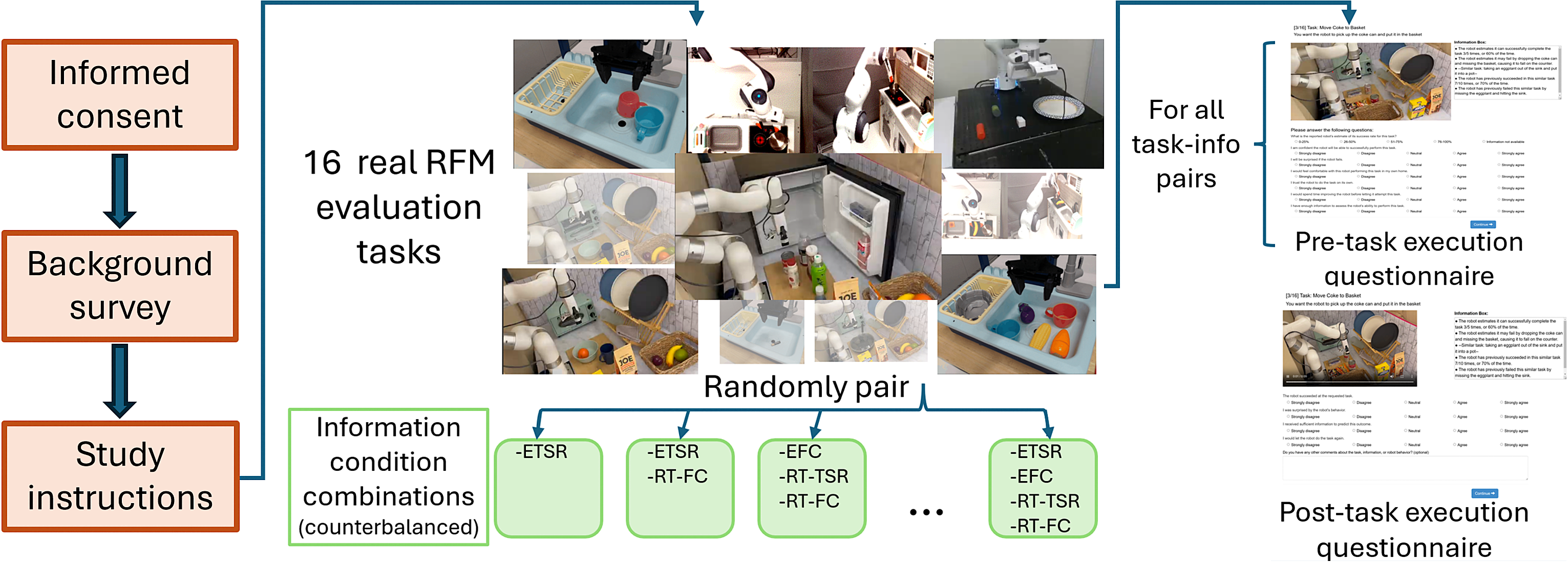}%
\vspace*{-3mm}
\caption{Overview of the study procedure. Users saw a successful or failed trajectory based on a probabilistic sample from the real evaluation success rate for that task. For each participant, the 16 information combinations where paired with the 16 tasks.}%
\vspace*{-6mm}
\label{fig:studyDiagram}
\end{figure*}

\textbf{Study Design}
\label{onlinestudy}
The within-subjects study had users report on their perception of a robot's ability to perform a task given different information types (Figure \ref{fig:studyDiagram}). We generated 16 information-task pairs for each participant. The information types were presented in all 16 permutations, including the case of no information, and randomly paired with the tasks. As RT-TSR and RT-FC could contain information about a task the user is yet to see, some participants saw information about a task before being presented with that task. To mitigate this and other potential ordering effects, we counterbalanced the information permutations using Latin square methodology. After participants provided informed consent, they were told that they would be making judgments about ``whether or not you want a robot to perform a task in your home.'' Each information type was explained and an example task was shown. Then, users filled out a background survey that asked for information such as age, occupation, and level of robot experience; and included the Personal Level Positive Attitude and Negative Attitude sub-scales of the General Attitudes Towards Robots Scale (GAToRS) survey [\citep{koverola_general_2022}. 

For each task, participants were shown a task request (e.g., “move the can of soup to the refrigerator”), a still image of the robot before execution, and an information box with the provided information. They then answered 5-point Likert questions about their perceptions of the robot, predictions of its behavior, and whether the information was sufficient to assess task performance; participants also reported the ETSR from the information box as a manipulation check. Following these questions, they were shown a video of the robot performing the task. Whether a success or failure was shown was determined by sampling the actual TSR from the collected data as to realistically reflect interactions with the model. Then, they were asked questions about their level of surprise and to reflect on the sufficiency of the provided information. They were also asked to judge if the task was completed successfully. The videos were slowed towards the end so it was easier to visualize the effects of the robot's actions; the video could also be rewound at any time. Before seeing the 16 tasks, users responded to questions in a practice task and the data from this practice was discarded. After seeing all tasks, they completed a post-study questionnaire.

The study questions, fully listed in the appendix, explore how information types influence user perceptions of a robot’s capabilities and willingness to use it. They were developed by partially following recommendations from [\citep{schrum_four_2020} for developing Likert questionnaires for HRI (e.g. a 5-point scale was intentionally chosen to mitigate fatigue due to the repetitiveness of the study). Participants were compensated \$9.07 for the $\sim$34 minute study. This procedure was approved by the Tufts University Institutional Review Board (IRB).

\begin{figure*}[t]
\centering
\captionsetup{justification=centering}
\includegraphics[width=.47\textwidth]{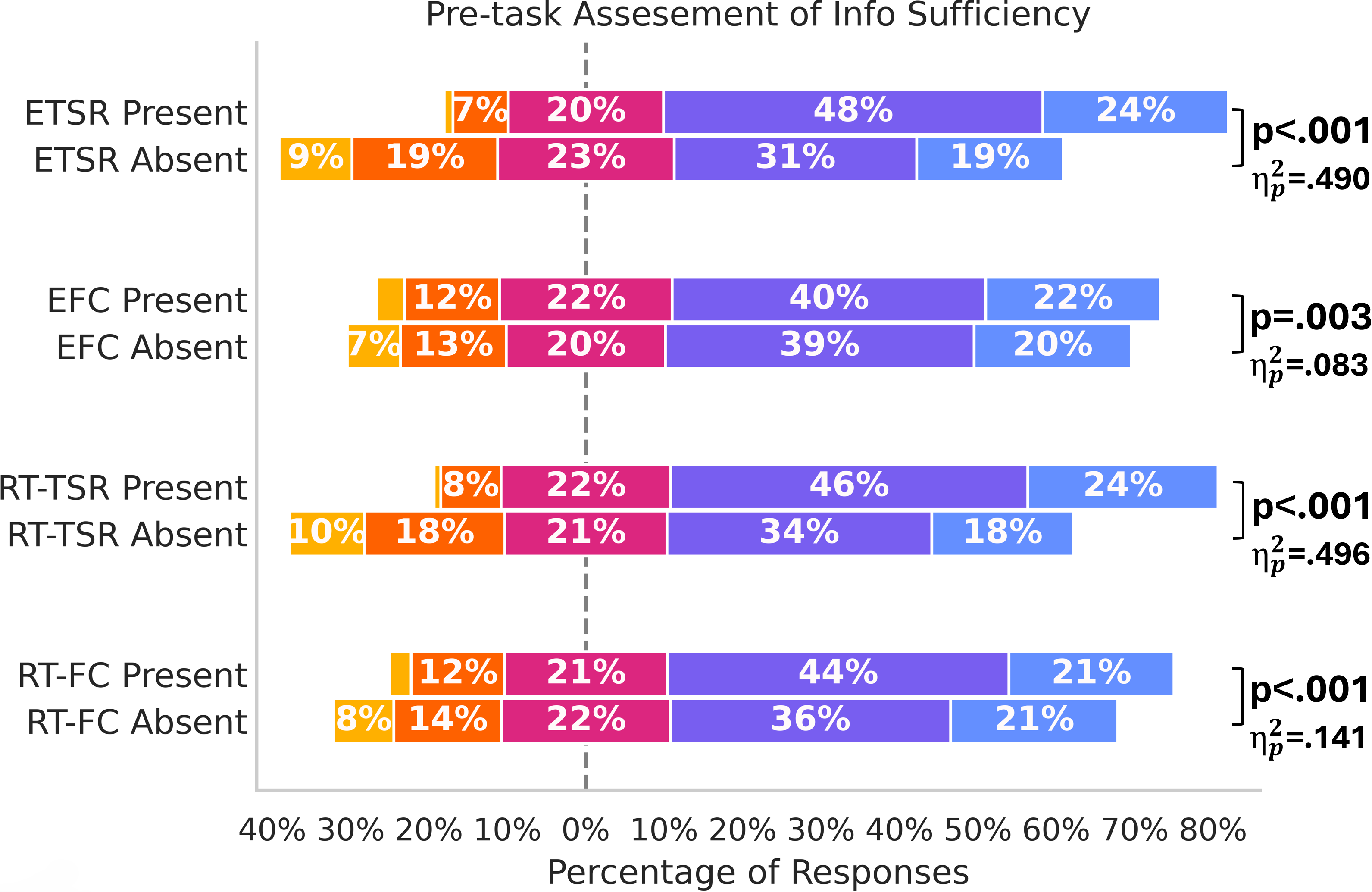}%
\hfill 
\includegraphics[width=.47\textwidth]{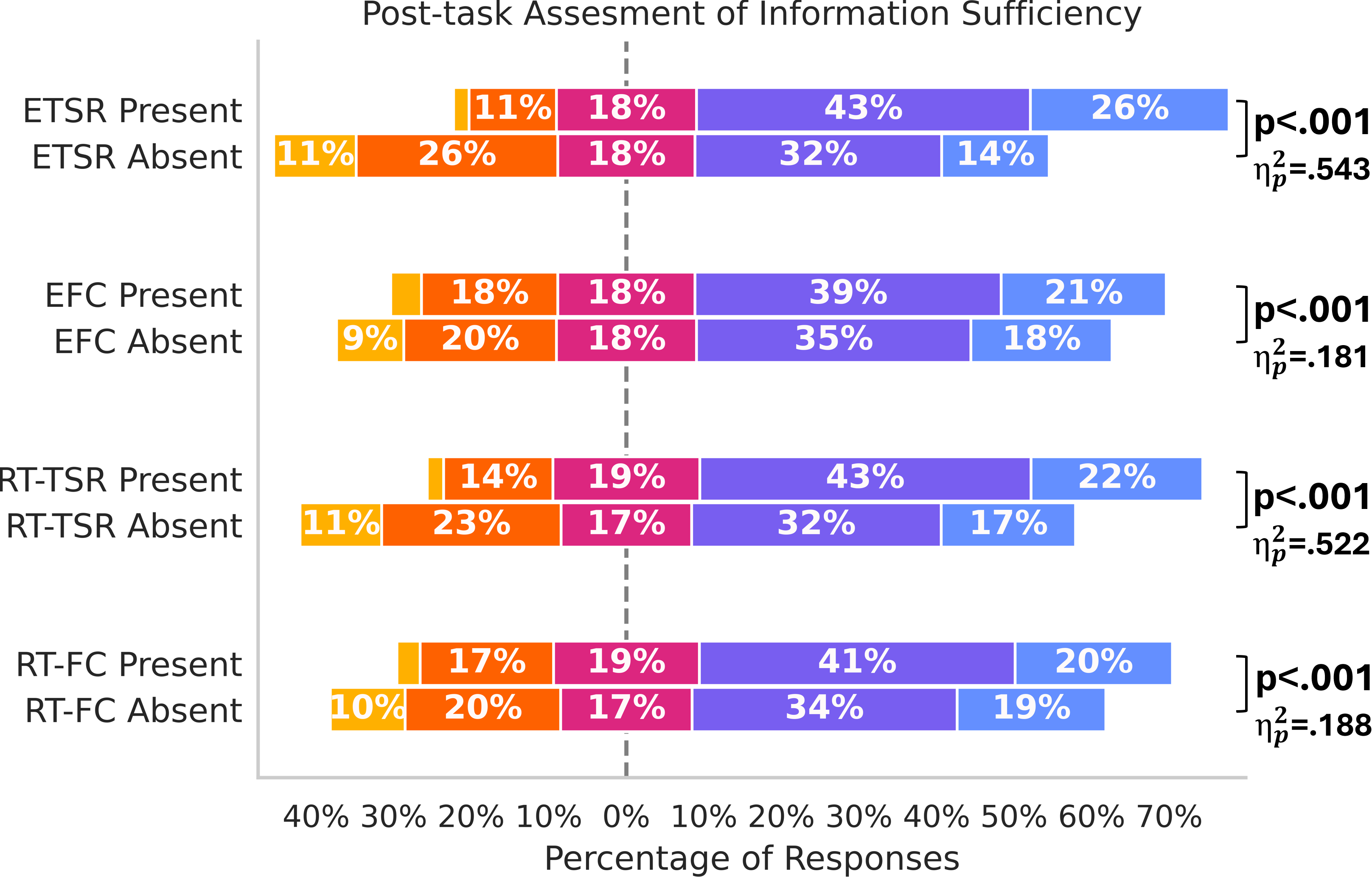}
\vspace*{-3.5mm}
\caption{Responses to the pre-task and post-task Likert questions of information sufficiency under different conditions. p-values and $\eta_p^2$ effect sizes from an RM-ANOVA are shown. \\
\textit{Legend}: \legendsquare{strongDisColor} Strongly Disagree/1, \legendsquare{disColor} Disagree/2, \legendsquare{neuColor} Neutral/3,  \legendsquare{agreeColor} Agree/4, \legendsquare{strongAgreeColor} Strongly Agree/5}%
\vspace*{-6mm}
\label{fig:info_anova}
\end{figure*}

\subsection{Results}
We recruited 112 participants from Prolific; 11 were excluded for not completing the study or consistently failing manipulation checks. Common demographic information of included participants can be found in the appendix. For statistical analysis, we used parametric methods, including RM-ANOVA where the presence or absence of an information type is treated as a repeated measure (resulting in 16 levels), and non-parametric methods including the Wilcoxon singed-rank test. Despite likert-item data being non-parametric, we used an RM-ANOVA specifically to account for potential interaction effects between the different information types. We use parametric tests when directly comparing the different information types. We used Bayesian tests [\citep{van_doorn_jasp_2021} for tests other than RM-ANOVA due to computate limitations.  

\textbf{Quantitative}
To investigate the impact of each information type on user perceptions and responses, we developed three hypotheses to test:
\begin{itemize}
    \item \textbf{H1} is that participants' perception of having sufficient information to assess the robot will improve with each additional information type, both before and after the robot attempts the task. This hypothesis is motivated by prior research in HRI related to trust that demonstrate that users want information and explanations about a robot’s behavior, especially regarding failure [\citep{wachowiak_when_2024, das_explainable_2021, riveiro_challenges_2022}.
    
    \item \textbf{H2} is that participants will report greater comfort with the robot performing the task as more information is available. This hypothesis is grounded in prior research that has shown transparency may increase trust and comfort in interaction, especially when related to safety [\citep{akalin_you_2022, fischer_increasing_2018}.
    
    \item  \textbf{H3} is that participants will find all information types useful, with ETSR being considered the most useful and equality among others. This hypothesis is more exploratory but central to understanding what information types should be made available to users. We expect ETSR to be reported as most useful since it is directly connected to whether the robot fails or succeeds in a binary sense.
\end{itemize}

 To analyze H1, we used two Likert questions that attempted to measure the perceived sufficiency of each information type, where higher values indicate greater sufficiency: ``I have enough information to assess the robot's ability to perform this task'' (asked before the robot attempted the task), and ``I received sufficient information to predict this outcome'' (asked after the task attempt). With an RM-ANOVA, we find that the presence of each information type significantly increases the user's reported perceived information sufficiency. We find noticeably larger effect sizes in the presence or absence of success rates than in failure cases. The distribution of responses and with p-values with effect sizes are shown in Figure \ref{fig:info_anova}. We found a positive correlation between people’s reported info sufficiency and the amount of info that they were presented with (i.e. 0-4 info types): $r=.265, p<.001$, $BF>1000$. These findings support \textbf{H1}.

For \textbf{H2}, we consider two pre-execution Likert questions: ``I would feel comfortable with this robot performing this task in my own home'' and ``I trust the robot to do the task on its own.'' Both questions relate to the user willingness to let the robot attempt the task. The amount of information present was a weak predictor of comfort for the ``home'' and ``trust'' questions (via a correlation analysis, $R^2=.008$, $F=13.107$, $p<.001$, $BF=35.86$ and $R^2=.011$, $F=17.439$, $p<.001$, $BF=309.22$ respectively). An RM-ANOVA of ETSR and RT-TSR shows that if the robot has a high ETSR and a high RT-TSR, user comfort increases ($p<.001$ for all cases). In contrast, neither EFC nor RT-FC increased reported user comfort for either question. We also intuitively find the higher success rate, the higher reported comfort for the ``home'' and ``trust'' questions ($R^2=.26$, $F=32.460$, $p<.001$, $BF>1000$ and $R^2=.26$, $F=32.434$, $p<.001$, $BF>1000$ respectively). These results partially support \textbf{H2}, with TSR leading to greater reported comfort.

\begin{figure*}
\centering
\begin{minipage}{.45\textwidth}
  \centering
  \includegraphics[width=.99\linewidth]{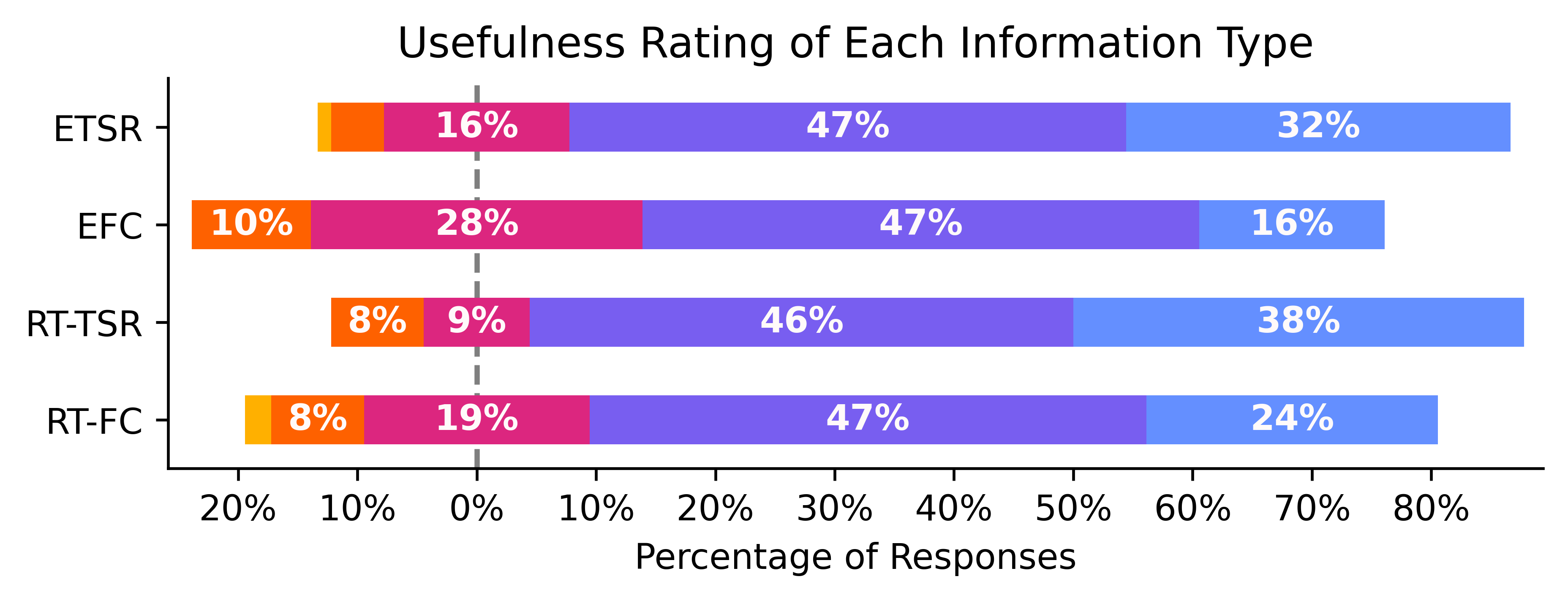}
  \vspace{-8mm}
  \captionof{figure}{Users in the online study generally reported \\ each information type as useful.}
  \label{fig:online_usefulness}
\end{minipage}%
\hspace{5mm}
\begin{minipage}{.45\textwidth}
  \centering
  \includegraphics[width=.99\linewidth]{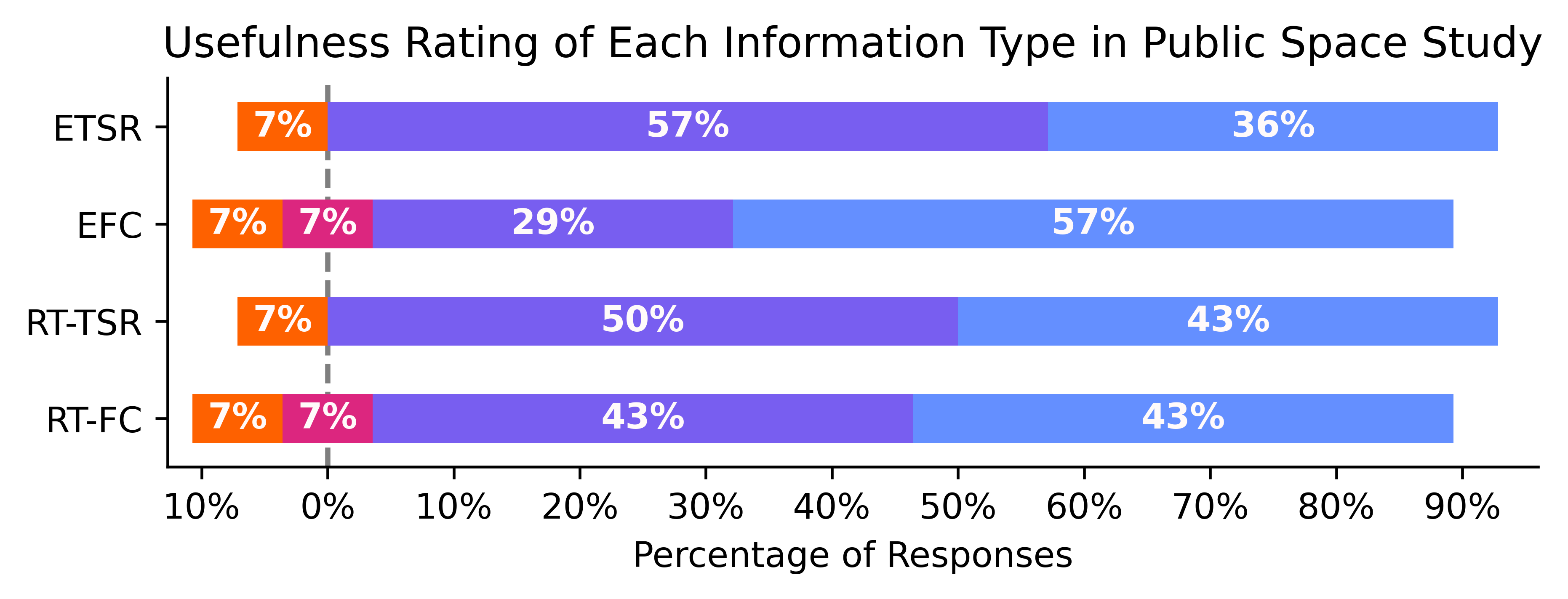}
  \vspace{-8mm}
  \captionof{figure}{Users in the follow-up study widely reported each information type as useful.}
  \label{fig:follow_up_usefulness}
\end{minipage}
\vspace{-5mm}
\end{figure*}

Subjective post-study user assessments of the utility of each information type are given in Figure \ref{fig:online_usefulness}. A large majority agreed or strongly agreed that all information types are useful (ETSR: 79\%, EFC: 63\%, RT-TSR: 84\%, RT-FC: 71\%). A Wilcoxon signed-rank test showed that ETSR was reported as more useful than EFC ($p<.001, BF=331.52$) and that RT-TSR was more useful than RT-FC ($p=.004, BF=12.46$). These results support the first half of \textbf{H3}, that users find all information types useful when evaluating robot performance. However, they do not support the latter part, as ETSR was not significantly more useful than all other types. 

Finally, although participants were not asked to make an explicit binary success/failure prediction, we treated responses to the pre-task statement “I am confident the robot will be able to successfully perform this task” as a proxy. “Agree” or “strongly agree” responses were coded as success predictions, and all others as failure. Prior work shows that change from Likert-scale responses to binary yields similar positive/negative patterns to binary-choice questions [\citep{suarez-garcia_exploring_2024, dolnicar_how_2007}. We conducted four independent-samples t-tests with information presence as the grouping variable and prediction accuracy as the dependent variable. Only ETSR significantly improved prediction accuracy ($p<.001, BF>1000$), with users correctly predicting outcomes 61.9\% of the time. While this does not capture nuanced expectations about robot behavior, it indicates that ETSR was most informative for binary success prediction.

\textbf{Qualitative}
To understand how users interpreted the different information types and what other information they may want, we asked two open-response questions in the post-study questionnaire: ``How did you use each information type to make your decisions? What made certain information more useful than other information?'' and ``What other information types may be useful to you to decide when a robot can or cannot reliably perform a task?'' The purpose of the latter question was to investigate both what users want and to identify potential gaps in user-wants and what is reported in RFM research. For analysis, we employed thematic inductive coding [\citep{ryan_techniques_2003, vanover_analyzing_2021}. For each question, a researcher came up with themes, codes within those themes, and a codebook with descriptions for each code. Two researchers then independently labeled each response with up to five codes with spreadsheet annotation. Disagreements were resolved through discussion. The final codebooks and code counts can be found in the appendix.

\textbf{Information Usage} From the responses to the information usage question, three themes emerged: preference, trust, and strategy. From those themes we developed 14 codes in total. We found a wide variety in user preferences for information types, with a divide between those who preferred real data and those who preferred estimates, although many participants relied on both types. 31 responses were tagged with the \textit{preferred real data} code and 30 tagged with \textit{preferred estimates}. Participants were also divided on whether they trusted estimates (18 explicitly reported to \textit{trust estimates} while 11 were \textit{skeptical of estimates}). Those who were skeptical typically also indicated that they preferred real data to make decisions: 
 \textbf{P. 45}, said real-data provides ``tangible evidence of its reliability'' and \textbf{P. 80}, said real-data is ``crucial because it provides real-world evidence of the robot’s capabilities and potential failures.'' In contrast, users who preferred estimates both mentioned trusting them and appreciated that ``it was a direct factor on whether or not the robot would accomplish the task.''

People often had specific strategies for using the information. \textit{Ordered preference} was tagged 20 times: users explicitly mentioned a ranking or process for using some information to support another. \textbf{P. 83} said, “I used success estimates to gauge the robot's reliability and past performance to assess trustworthiness.” Five of these participants mentioned using thresholds: \textbf{P. 8} noted that an ETSR “below 70\% felt iffy [unreliable],” and that RT-TSR “helped me with my confidence in the absence of a success rate for the main task itself.” Nineteen participants mentioned that their ability to use real-data depended on how similar the tasks were. \textbf{P. 48} for instance explained their usage of real-data ``would depend on how similar it was to the actual task; if it was not that similar then I would disregard the information and just use my gut.'' Overall, these responses highlight diverse user strategies and preferences and underscore the importance of quantifying task similarity in RFMs as an indicator of performance on new tasks.

\textbf{Participant Suggestions for Additional Information Types} We asked what other information would be useful when determining robot reliability. From the responses we derived seven themes: Robustness, Robot Capability, Learning, Failure, Task, and Miscellaneous. Within those themes we developed 18 codes in total. The most common request was for \textit{more information} like the information types already available, such as more estimated failure cases or more information about related tasks (38 participants). This is consistent with the quantitative analysis of the presence of each information type being useful. However, users also expressed diverse and rich desires about what other information they may want.

\textit{Robot capability} was the second most common information type requested. Participant expressed a desire to know about the robot's physical specifications, such as the ``robot’s strength and dexterity'' or its sensing capabilities, such as ``color identification and motion detection.'' Similarly, 8 participants mentioned that they wanted to know about \textit{speed} or how fast the robot could perform the task. Despite EFC and RT-FC being rated as less useful than their TSR counter part, many users identified other aspects of failure as important (the code \textit{failure rate and case} was tagged 23 times, the code \textit{failure degree} tagged 7, and the code \textit{failure recovery} tagged 5 times). \textit{Environment factors} and comments about robustness were also frequently mentioned. \textbf{P. 43} provided a detailed response that captured well the \textit{environment factors,} \textit{robustness to environment,} and \textit{robustness to task} codes: ``It would be helpful to know more about the robot’s past experiences with similar tasks in different environments or settings, as well as its ability to adapt to new challenges. Also, understanding how the robot handles unexpected obstacles or changes in the task would give a better idea of its reliability.''

The need for information about the robot's algorithm and real performance was also present, albeit less focused and with less consensus than topics such as robot capability. Intuitively, people wanted real success rate or failure cases on the requested task (13 participants), or wanted to watch the robot repeatedly attempt the task (8 participants). 31 participants wanted to know more about the nature of the task itself or ``thought about how easy or difficult it would be for the robot to perform the task.'' Some wanted information about the robot's ability to learn. \textbf{P. 65} for example said ``understanding how the robot learns from past mistakes would help me trust it more.''

\begin{figure}[t]
\centering
\includegraphics[width=.44\textwidth]{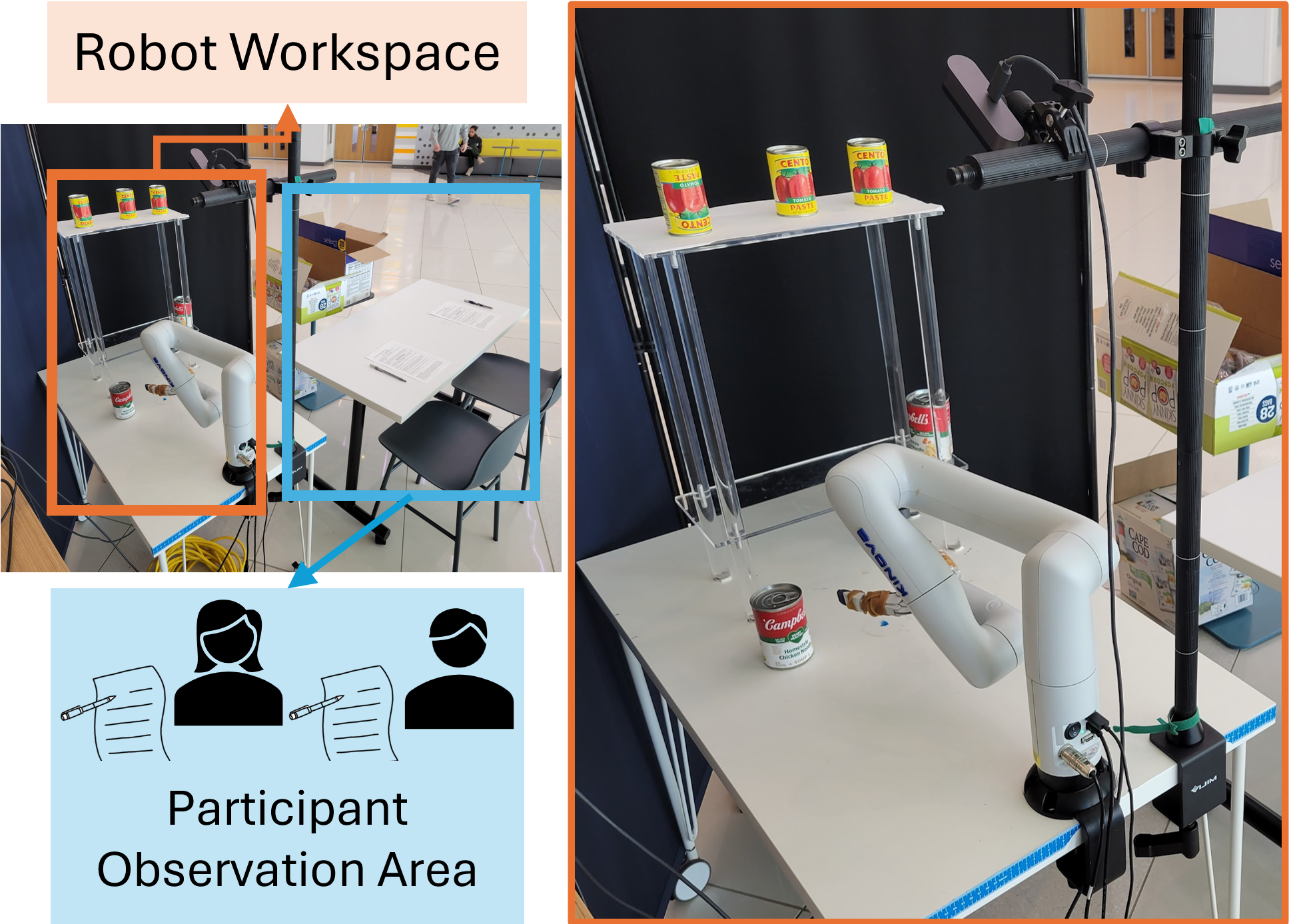}%
\caption{The in-person study was conducted in a University building lobby. The robot repeatedly attempted to put the cans on the shelf.}{}%
\label{fig:followUpStudyDiagram}
\end{figure}

\section{Exploring User Information Needs with an Embodied Robot (in-person study)}
In the online study, we identified user information preferences and needs with a variety of tasks and real evaluation data. In this in-person study, we sought to identify the effect of embodiment: checking for differences in the information types users perceived as valuable in-person compared to their preferences when evaluating videos and exploring what new information types they may want. Watching a robot perform a task in real-time offers the benefit of seeing how its behavior affects the environment, and may change the users' assessment of the robot from embodiment alone [\citep{seo_poor_2015}.

\subsection{Procedure}
The study was conducted using a Kinova Gen3 Lite robot in a University building lobby. To maintain similarity with the tasks from the online study, we designed a kitchen chore-inspired task in which the robot had to place cans of soup on a shelf. To enable the robot to autonomously complete and demonstrate the task, we trained the robot with Diffusion Policy (DP) [\citep{chi_diffusion_2024} using an implementation from LeRobot [\citep{cadene_lerobot_2024} and further adapted to ROS [\citep{stanford_artificial_intelligence_laboratory_et_al_robotic_2018}. While DP is not a RFM, it is nevertheless a start-of-the-art imitation learning algorithm often used as a baseline for RFMs and was feasible to train and run inference locally using a NVIDIA RTX-3080 (current work is trending towards making RFMs more affordable to use [\citep{wen_tinyvla_2024, belkhale_minivla_2024}). The policy completed the task most of the time but would also fail regularly:  by failing to grasp the can (hovering over it or pushing its gripper into it), pushing the can into the shelf itself instead of placing it, or dropping the can onto the table. Participants observed the robot repeatedly attempting the task, with a researcher resetting the environment after each run. The robot served as an example for answering questions. Participants were recruited from passersby and with posters. After providing consent, they were given a questionnaire which described the task that the robot was doing, an information box that explained each information type, and provided examples of each type for the robot's can-clean-up task. Users were asked to think about how the provided information may be helpful or not when deciding whether a robot is capable enough to perform tasks in a home. 

Like the online study, they were asked a Likert question for indicating how useful each information type is for determining whether one would want to use the robot. We also asked two open-ended questions: ``What other information types may be useful to you to decide when a robot can or cannot reliably perform a task?'' (also in the online study); and ``How would you decide whether or not the robot was good enough at a task to want [to use] one?'' Finally, they offered a variety of snacks as compensation. The study took five to ten minutes to complete and was approved by the Tufts University IRB.   

\subsection{Results}
\textbf{Participants} In total, 14 people participated in the study. 

\textbf{Quantitative} The Likert response data can be found in Figure \ref{fig:follow_up_usefulness}. Like in the online study, a large majority of users reported all information types as being useful. Unlike in the online study, we did not find any significant differences between information types. 

\textbf{Qualitative} To analyze the open-ended questions about other information types, we used the same codebook developed from the online study. Users had similar preferences as the online study about what information they would find useful. While only 2 participants wanted more information, \textit{robot capability} was the most frequently tagged code. Multiple participants brought up things like space utilization and workspace, which were not brought up in the online study. \textbf{P. A12} for example said ``At what speed can tasks be performed? (e.g. tx/min) Cost of automation? space utilization, safety, types of products.'' And \textbf{P. A10} wrote ``How much installation the robot would require or how much space it takes up. Also how long it takes to perform the task.'' Concerns about failures were also more personal or pronounced. \textbf{P. A4} wanted to know if ``In the case of failure will it damage anything or anyone,'' and \textbf{P. A5} expressed concern ``if it drops it [the soup can] on my ceramic tile floor and doing damage.'' Responses to the second open response question generally reported wanting a highly capable and safe robot, with three participants specifying the robot should be roughly as capable as a human at the task. These results corroborate many of the findings from the online study, and provide insights into the information needs of users who are physically co-present with the robot.

\section{Discussion}
\textbf{Limitations and Future Work} 
This study has several limitations regarding design and generalizability. Most tasks involved a stand-alone robot arm and kitchen-inspired chores; while common in RFM evaluation, these settings may not generalize to other embodiments (e.g., quadrupeds) or task types (e.g., safety-critical tasks). Nonetheless, we expect many findings to transfer across morphologies, even if absolute trust and comfort levels shift [\citep{kunold_not_2023}, as users may value similar information types. Performance information was presented only as text, leaving the effects of alternative presentations on user expectations and decision-making as future work. Finally, while we designed the study to isolate the effects of performance information, future work should examine its role in collaborative settings, such as how users decide whether to teach a robot or allocate tasks.

\textbf{Generalization to other Robot Models}
A key contribution of this work is not only the behavior of the RFM models but also the use of representative RFM research tasks and success criteria. This work serves to inform RFM evaluations and deployments. However, given that the task videos were pre-recorded and users were not explicitly told or given information that the robot's where using an RFM, the results from the study should equally apply to non-RFM-based robot systems, for example robots trained with single-task imitation learning. Future work should study and evaluate RFMs when the user can freely request an arbitrary, novel-task task and interact in real time with a robot.  

\textbf{Implications for Future RFM Research} 
Our results show that people generally interpret TSR as intended: higher TSR increases confidence that a robot will succeed. However, TSR alone is insufficient for informing users about performance on novel tasks. We therefore recommend that RFM evaluations routinely report failure cases, including both worst-case and common failures, as this information benefits novices and experts alike. While some metrics (e.g., task speed) are easy to report, others are harder to formalize; for example, robustness to static environments can be partially evaluated in simulation [\citep{ wang_roboeval_2025}, robustness to dynamic interference remain difficult to quantify. Our findings also underscore the importance of both real data and estimates: real data require principled measures of task similarity, and novel tasks require reliable performance estimation. Recent work has expanded simulation-based RFM evaluation to include behavioral metrics, such as motion smoothness, and improved failure detection [\citep{wang_roboeval_2025, xu_can_2025}. As expert evaluations grow more holistic, our results highlight the need for this information to be made available to end-users.

Future work is needed to understand how people use TSR to wholistically predict robot behavior, as user expectations may not align with actual behavior even when accurate metrics are provided. Although TSR strongly correlates with trust and confidence, it remains unclear if users are well calibrated to a robot’s true performance over time, particularly for tasks performed repeatedly. As people gain experience with a robot performing a certain task, the value of certain information may change. In the online study, when participants reported the highest confidence in the robot's ability to successfully perform the task, the average ETSR was only $\sim$76\%, suggesting a potential miscalibration between a user's estimate of success likelihood and TSRs [\citep{xu_can_2025}. Like other robotic systems [\citep{das_explainable_2021}, RFMs must explain failures in ways that avoid over- or under-trust.

\section{Conclusion}
RFMs have the potential to enable robots to perform a wide range of tasks for users, especially in home environments. Through user studies, this work begins to address the critical need to understand how end users interpret performance information about these models. Based on our findings, we further provided recommendations to make future RFM evaluations and RFM deployments more user-friendly and interpretable. 

\bibliography{references_ijcai}
\newpage

\appendix
\onecolumn
\vspace{-8mm}
\section{Task list}
Here is a list of all participant facing tasks used in the study. In the study, the task description was appended to the prefix ``You want the robot to.'' The failure case (FC) description was appended to the prefix ``The robot estimates it may fail by'' for EFC and ``The robot has previously failed this similar task by'' for RTFC. The TSR information was appended to the prefix ``The robot estimates it can successfully complete the task'' for ETSR and ``The robot has previously succeeded in this similar task'' for RT-TSR. 

\begin{table}[h!]
\centering
\begin{tabular}{| m{1.8cm} | m{3cm} | m{4cm} | m{6.2cm}|}
\hline
\textbf{Task \newline (Model)} & \textbf{Task Image} & \textbf{Task Description} & \textbf{Task Information and Similar Task} \\ 
\hline
Stack cups \newline(OpenVLA)& \includegraphics[width=3cm]{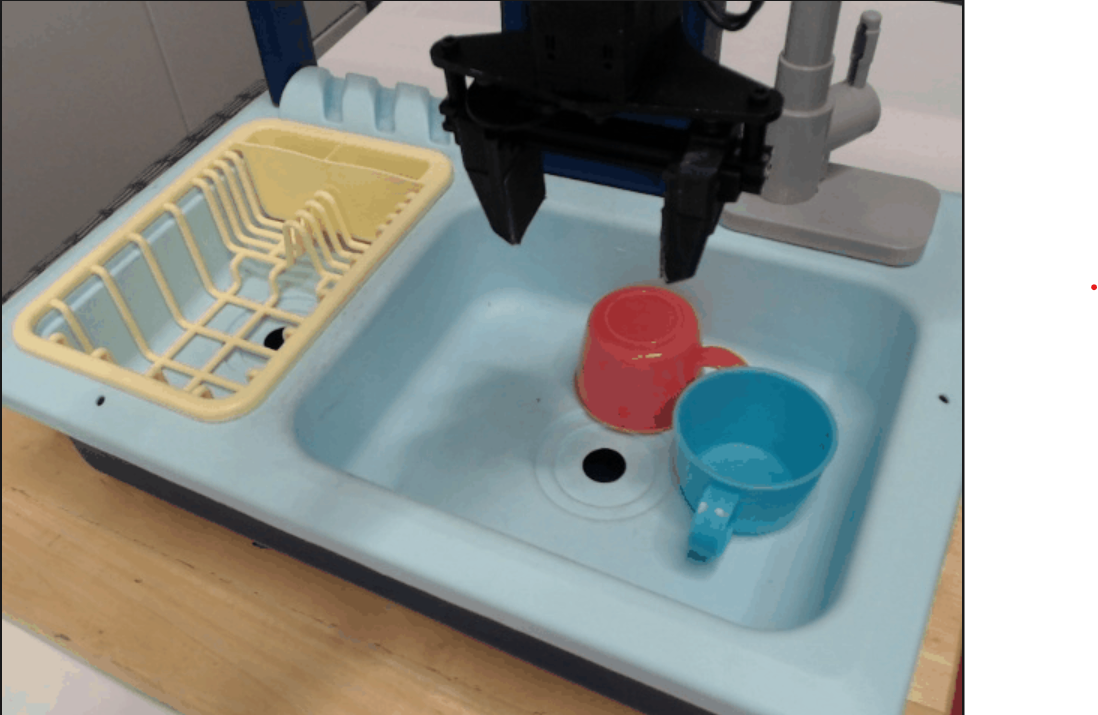} & Stack the blue cup face-up on top of the upside down pink cup. & \textbf{TSR:} 4/10 times, or 40\% \newline
\textbf{FC}: Picking up the cup but dropping it in the sink, missing the pink cup.\newline
\textbf{Similar Task:} Lift pan lid\\

\hline
 Put away \newline soup \newline(Baku)& \includegraphics[width=3cm]{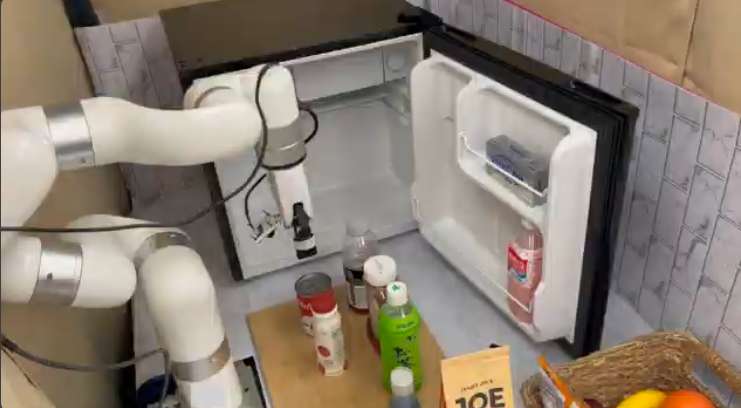} & Move the can of soup to the refrigerator.  & \textbf{TSR}:4/5 times, or 80\%  \newline
\textbf{FC}: Grasping the can but then halting, failing to put it in the refrigerator.\newline
\textbf{Similar Task:} Put away bottle \\
\hline

Close oven \newline door \newline(Baku)& \includegraphics[width=3cm]{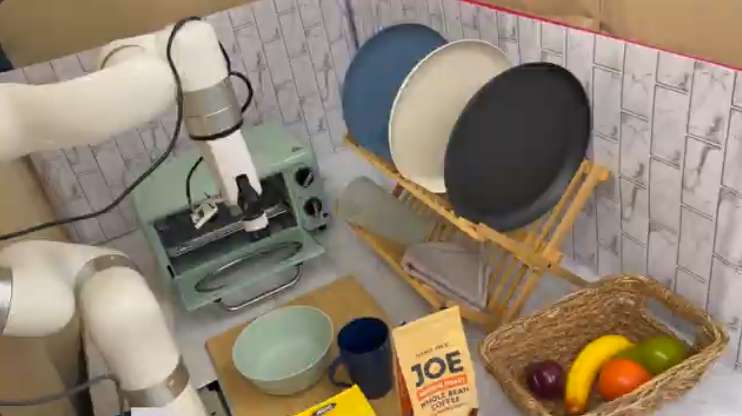} & Close the oven door.  & \textbf{TSR:}  4/5 times, or 80\% \newline
\textbf{FC}: Not grasping the handle, leaving the oven door open. \newline
\textbf{Similar Task:} Make toast \\
\hline

 Remove battery \newline(OpenVLA)& \includegraphics[width=3cm]{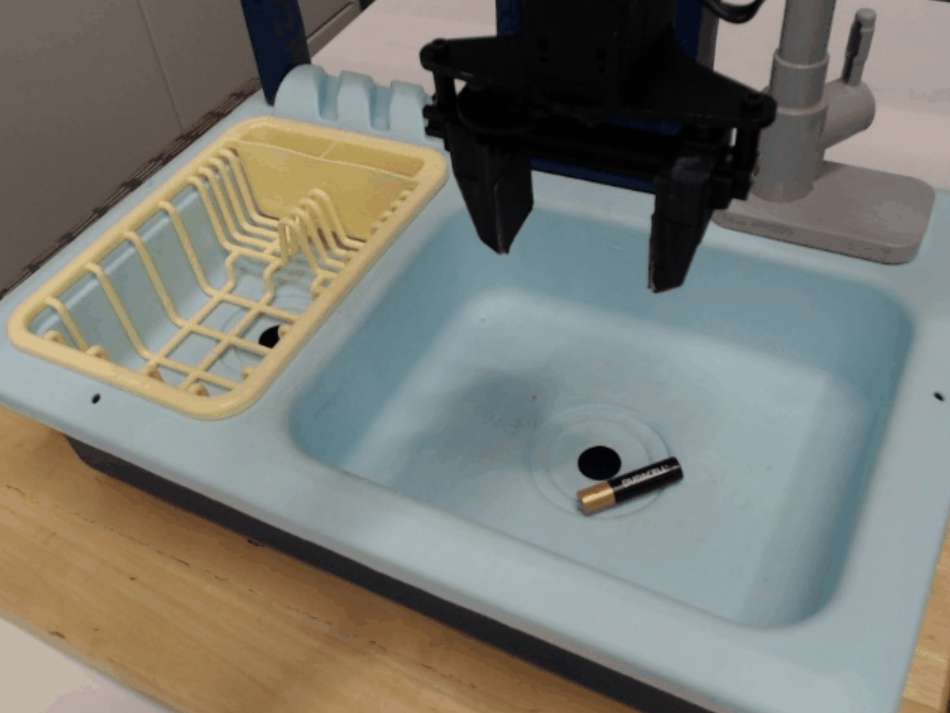} & Lift and remove the battery from sink. & \textbf{TSR:} 7/10 times, or 70\% \newline
\textbf{FC}: Not grasping the battery and then pushes the battery around in the sink.\newline
\textbf{Similar Task:} Lift pan lid \\
\hline

Move salt \newline(OpenVLA)& \includegraphics[width=3cm]{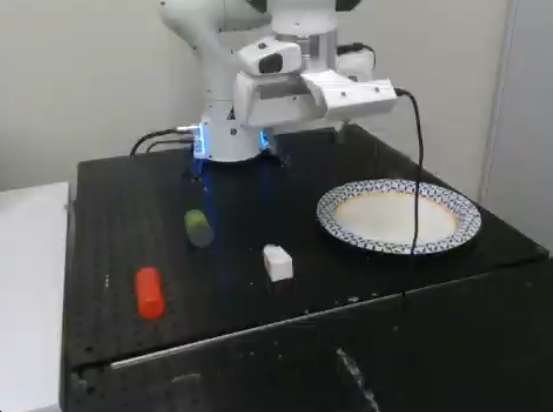} & Move the white salt shaker onto the plate. & \textbf{TSR:} 9/12 times, or 75\%  \newline
\textbf{FC}: Trying to pick up the wrong object. \newline
\textbf{Similar Task:} Put carrot on plate\\
\hline

 Lift \newline orange \newline(Baku)& \includegraphics[width=3cm]{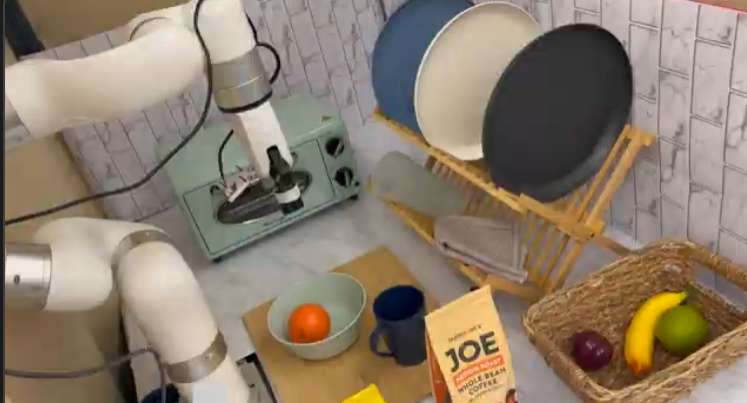} & Lift up the orange out of the bowl. & \textbf{TSR:} 4/5 times, or 80\% \newline
\textbf{FC}: Not grasping the orange, leaving it in the bowl.\newline
\textbf{Similar Task:} Remove battery \\
\hline

Put \newline eggplant \newline in pot \newline(OpenVLA)& \includegraphics[width=3cm]{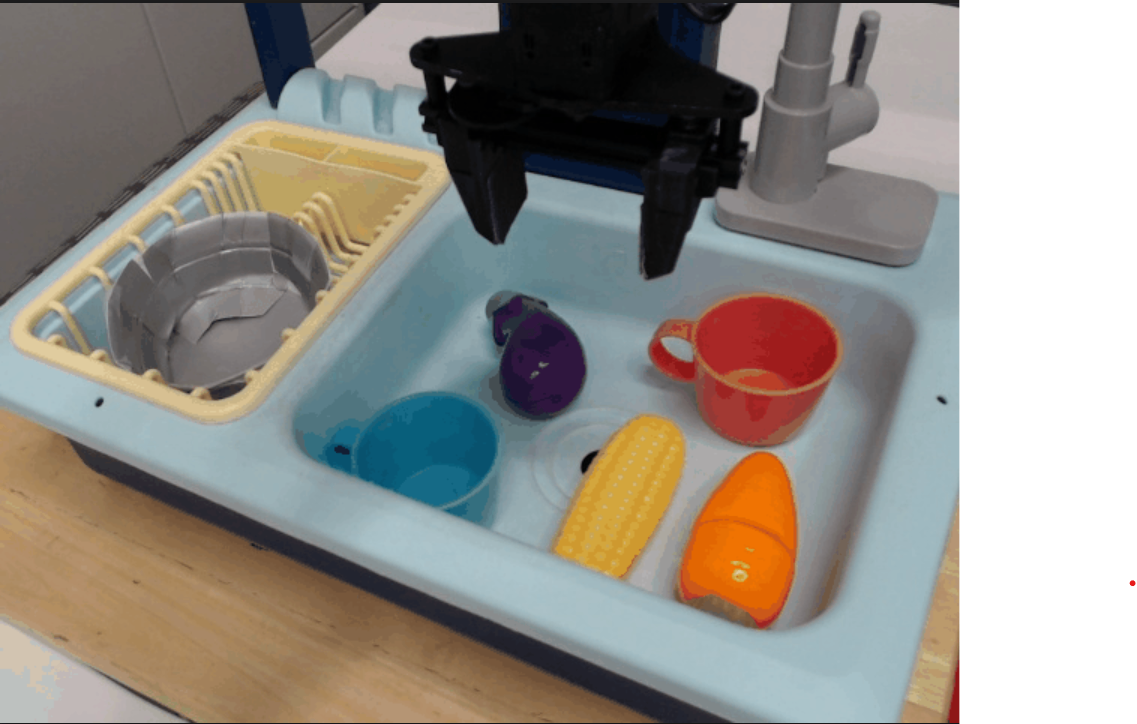} &  Take the eggplant out of the sink and put it into the pot. & \textbf{TSR:} 7/10 times, or 70\% \newline
\textbf{FC}: Missing the eggplant and hitting the sink instead. \newline
\textbf{Similar Task:} Put coke in basket\\
\hline

\hline
\end{tabular}
\caption{List of participant facing tasks used in the online study.}
\label{table:task_table}
\end{table}

\newpage

\begin{table}[h!]
\centering
\begin{tabular}{| m{1.8cm} | m{3cm} | m{4cm} | m{6.2cm}|}
\hline
\textbf{Task} & \textbf{Task Image} & \textbf{Task Description} & \textbf{Task Information and Similar Task} \\ 

\hline
 Lift pan lid \newline(Baku)& \includegraphics[width=3cm]{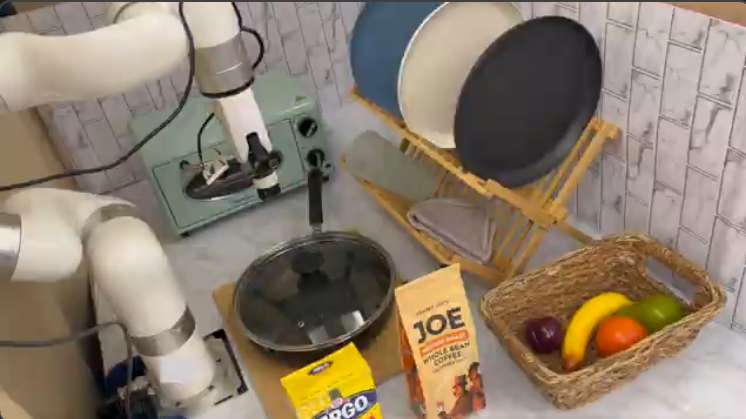} & Lift up the lid of the pan off the pan. & \textbf{TSR:} 4/5 times, or 80\%  \newline
\textbf{FC}: Partly grasping and then dropping the pan lid handle, causing the pan to shake as the lid falls back on top.\newline
\textbf{Similar Task:} Remove battery\\
\hline

 Move \newline banana \newline to sink \newline(MDT)& \includegraphics[width=3cm]{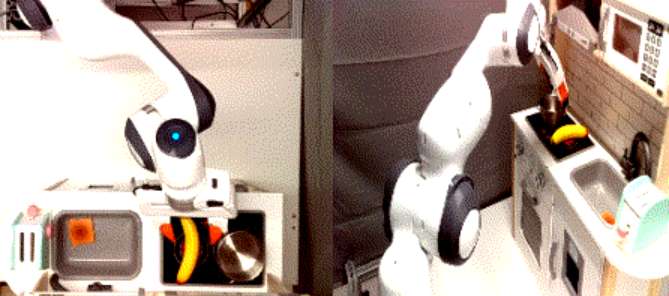} & Move the banana from the stove and place it into the sink. & \textbf{TSR:} 4/5 times, or 80\% \newline
\textbf{FC}: Not picking up the banana, putting nothing into the sink.\newline
\textbf{Similar Task:} Move banana to stove \\
\hline

 Move \newline banana \newline to stove \newline(MDT)& \includegraphics[width=3cm]{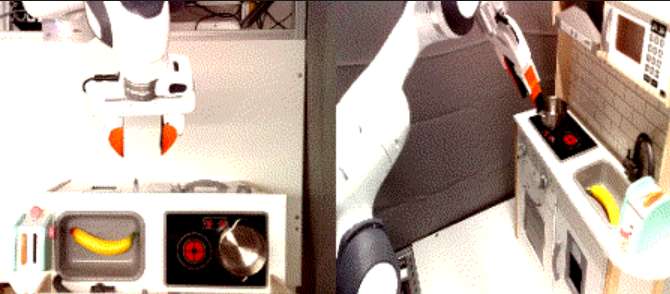} &  Move the banana from the sink to the stove. & \textbf{TSR:} 4/5 times, or 80\% \newline
\textbf{FC}: Being unable to pick up the banana from the sink at the correct location and may push it's gripper into the sink.\newline
\textbf{Similar Task:} Move banana to sink \\
\hline

\hline
 Move pot \newline to sink \newline(MDT)& \includegraphics[width=3cm]{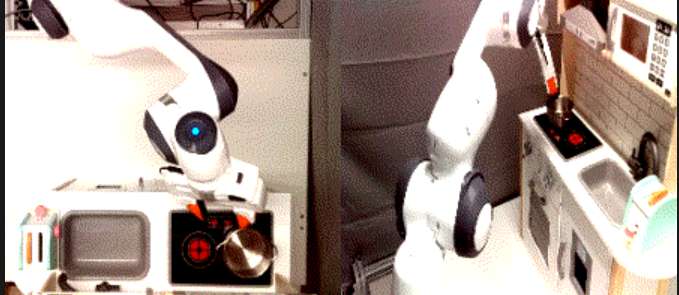} & Move the pot from the stove to the sink & \textbf{TSR:} 4/5 times, or 80\%  \newline
\textbf{FC}: Being unable to grasp the pot, moving it around on the stove and towards the edge. \newline
\textbf{Similar Task:} Move banana to sink\\

\hline
 Put away \newline coke \newline(Baku)& \includegraphics[width=3cm]{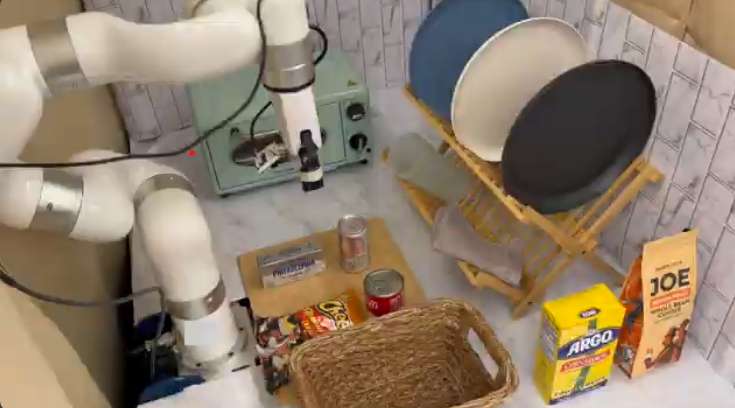} & Pick up the coke can and put it in the basket. & \textbf{TSR:} 3/5 times, or 60\% \newline
\textbf{FC}: Dropping the coke can and missing the basket, causing it to fall on the counter.\newline
\textbf{Similar Task:} Put eggplant in pot\\
\hline

Put \newline carrot \newline on plate \newline(OpenVLA)& \includegraphics[width=3cm]{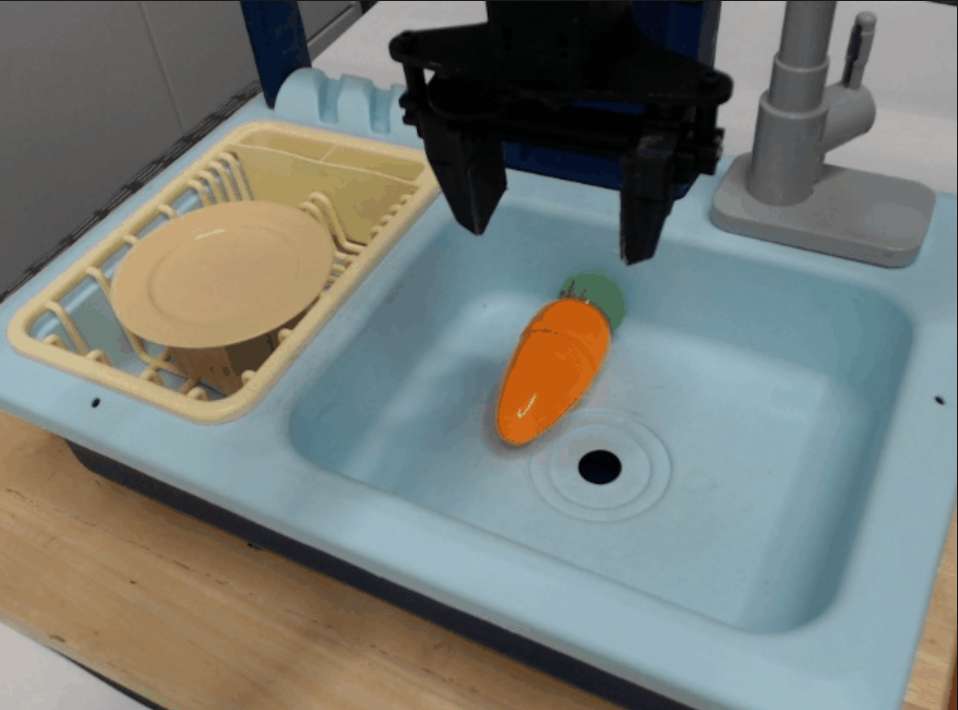} & Pick up the carrot from the sink and put it on the plate. & \textbf{TSR:} 4/10 times, or 40\% \newline
\textbf{FC}: Being unable to grasp the carrot and/or by knocking over the plate. \newline
\textbf{Similar Task:} Move salt\\
\hline

 Wipe board \newline(Baku)& \includegraphics[width=3cm]{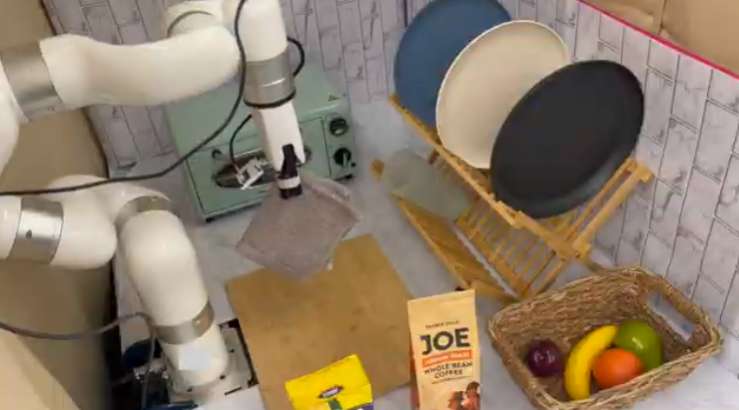} & Use the towel to wipe down the cutting board. & \textbf{TSR:} 5/5 times, or 100\% \newline
\textbf{FC}: Not available. \newline
\textbf{Similar Task:} Make toast \\
\hline

 Make toast \newline(MDT)& \includegraphics[width=3cm]{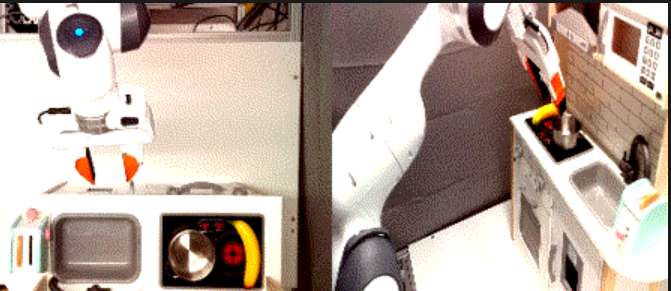} & Push down the toaster lever to make toast & \textbf{TSR:} 1/5 times, or 20\% \newline
\textbf{FC}: Not being able to push down the lever.\newline
\textbf{Similar Task:} Close oven door \\
\hline

 Put away \newline bottle \newline(Baku)& \includegraphics[width=3cm]{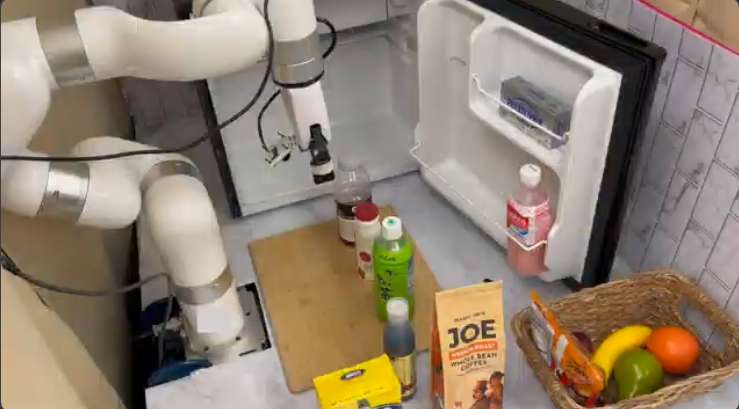} & Move the green bottle of tea to the refrigerator door. & \textbf{TSR:} 1/5 times, or 20\% \newline
\textbf{FC}: Attempting to pick up an object that is not the green tea.\newline
\textbf{Similar Task:} Put away soup\\
\hline

 (example task) \newline Put away \newline ball \newline(TinyVLA)& \includegraphics[width=3cm]{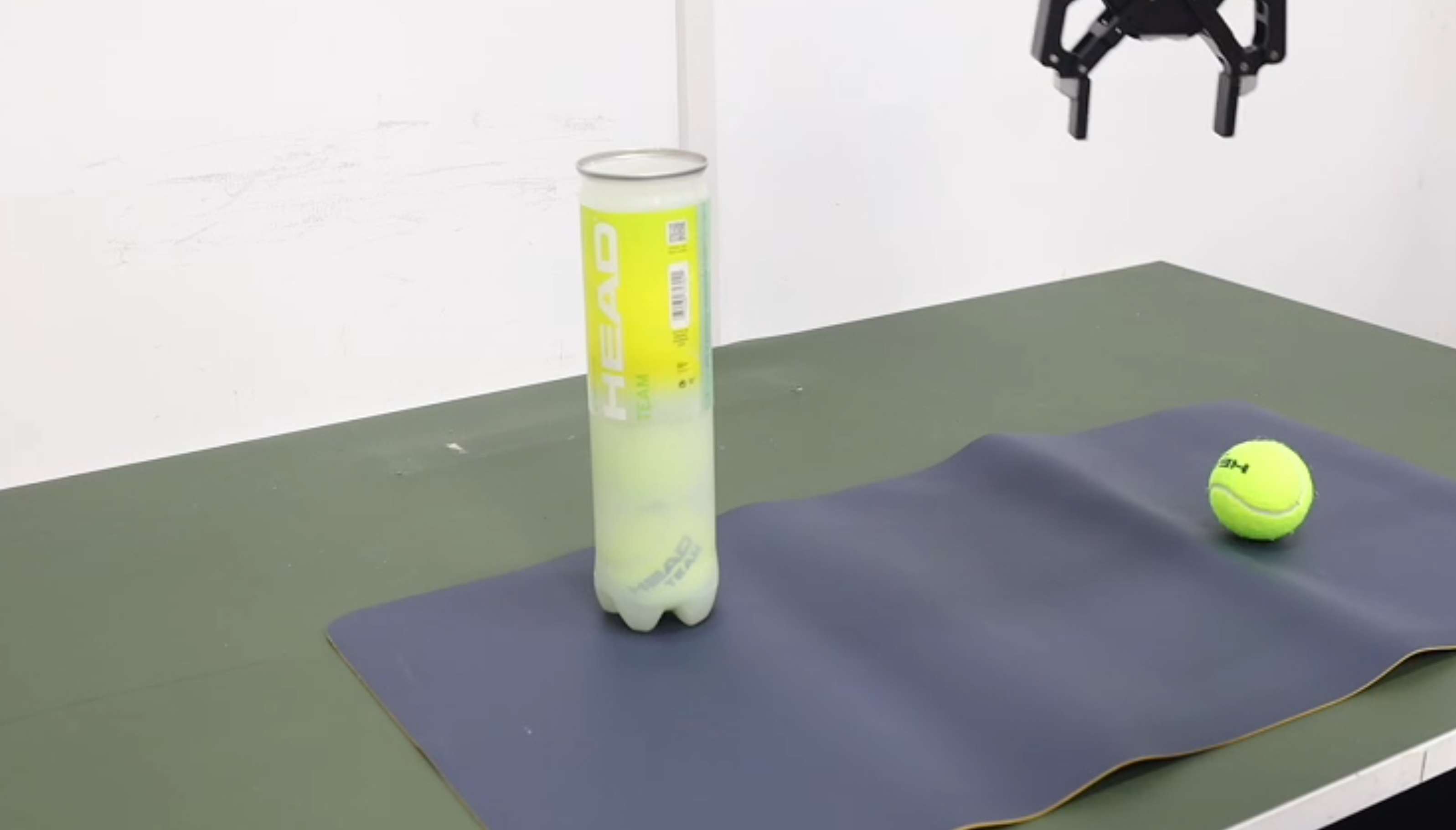} &  Put the tennis ball in the tube. & \textbf{TSR:} 4/5 times, or 80\% \newline
\textbf{FC}: Knocking over the tube.  \newline
\textbf{Similar Task:} (example) Put carrot in bowl \newline \textbf{RT-TSR}: 3/10 times, or 30\% \newline \textbf{RT-FC}: Being unable to the grasp the carrot. \\
\hline
\end{tabular}
\caption{List of participant facing tasks used in the online study (continued).}
\label{table:task_table}
\end{table}

\newpage

\section{Codebooks}
\begin{table}[h!]
\centering
\begin{tabular}{|m{2cm}|m{4cm}|m{10cm}|}
\hline
\textbf{Theme} & \textbf{Code} & \textbf{Description} \\
\hline
Trust & Trusted estimates & Someone who explicitly indicated that they trusted the estimates or thought the estimates could be relied on \\
\hline
Trust & Skeptical of estimates & Someone who expressed skepticism over if the estimates could be relied on or not/their value. \\
\hline
Trust & Performance threshold & Someone who mentioned a numeric threshold or some threshold of reported performance to where they could make decisions about the task. \\
\hline
Strategy & All info & Mentions all the information was useful \\
\hline
Strategy & Ordered preference & Mentions how they relied one or more types of info but then also used the others as backup or support \\
\hline
Strategy & Mainly used estimated TSR & Someone who more or less just says they used estimated TSR only \\
\hline
Strategy & Past performance & Someone who says they used the previous tasks or data to help them decide (implies sort of meta-learning the study) \\
\hline
Strategy & Similarity & Someone who mentions the value of the real data was based on how similar the task was \\
\hline
Strategy & Risk calculation & Someone who mentions trying to estimate failure risk \\
\hline
Strategy & Compared real and estimates & Explicitly compares real and estimates \\
\hline
Preference & Preferred estimates & Explicitly mentions they preferred estimates \\
\hline
Preference & Preferred real data & Explicitly mentions they preferred real data \\
\hline
Preference & Preferred success rates & Explicitly mentions they preferred success rates \\
\hline
Preference & Preferred failures & Explicitly mentions they preferred failures \\
\hline
\end{tabular}
\caption{Thematic codebook developed for the responses to the post-study open-response questions: ``How did you use each type of information to make your decision? What made certain information more useful than other information?''}
\end{table}

\begin{table}[h!]
\centering
\begin{tabular}{|m{2cm}|m{2cm}|m{10cm}|}
\hline
\textbf{Theme} & \textbf{Code} & \textbf{Description} \\
\hline
Robot \newline capability & Robot \newline capability & Wanting to know more about a robot's general specifications, sensing capabilities, or algorithmic ability. \\
\hline
Robot \newline capability & Speed & Wanting to know how fast the robot can perform the task. \\
\hline
Failure & Failure \newline degree & Wanting to know about how bad a failure could be/about the range of possible failures. \\
\hline
Failure & Failure rate \newline and cases & Wanting to know about failure rate as opposed to success rate; wanting to know what in the environment could be damaged or affected because of the robot executing the task; failures unrelated to the requested task. \\
\hline
Robustness & Environment factors & Wanting to know more about the environment: its dynamics and how it may affect the robot’s performance. \\
\hline
Robustness & Robustness to task & Wanting to know about how robust the robot is to minor variations in the requested task. \\
\hline
Robustness & Robustness to \newline environment & Wanting to know about how robust the robot is to external or environmental factors separate from the task. \\
\hline
Robustness & Failure \newline recovery & How capable the robot is at recovering from failures \\
\hline
Task & Task \newline difficulty & Wanting some sort of measure or assessment of task difficulty or complexity; wanting to know something about how good the robot is relative to human performance on the task.  \\
\hline
Task & Task type & General info about the nature of the task; wanting to know the horizon of the task, as in if it is a short or long task/single step vs multi-step. \\
\hline
Task & Performance on requested task & Wanting to know about how good the robot is on the actual task (not like estimated success rate, but like actual performance). \\
\hline
Task & Watch \newline repeated \newline attempts & Expressing a desire to watch the robot attempt the task one or more times. \\
\hline
Learning & Ability to\newline learn & Wanting to know how well the robot can adapt/learn as it executes a task or attempts a task multiple times. \\
\hline
Learning & Learning \newline experience & Wanting to know how the robot was trained or how it previously learned to do the task. \\
\hline
Miscellaneous & Human \newline collaboration & Wanting to know how well the robot could collaborate with a human or handle a human intervention; wanting to know about the robots ability to learn from people. \\
\hline
Miscellaneous & Satisfied & The participant does not express a desire for more information or is already satisfied with the provided information. \\
\hline
Miscellaneous & More info & Expressing a desire for more of the four information types already provided. \\
\hline
Miscellaneous & Deployment duration & Wanting to know about how long the robot has been deployed for or its deployment history. \\
\hline
\end{tabular}
\caption{Thematic codebook developed for the responses to the post-study open-response question: ``What other types of information may be useful to you to decide when a robot can or cannot reliably perform a task?''}
\end{table}

\clearpage
\section{Code Counts}

\begin{figure*}[h]
\centering
\includegraphics[width=.99\textwidth]{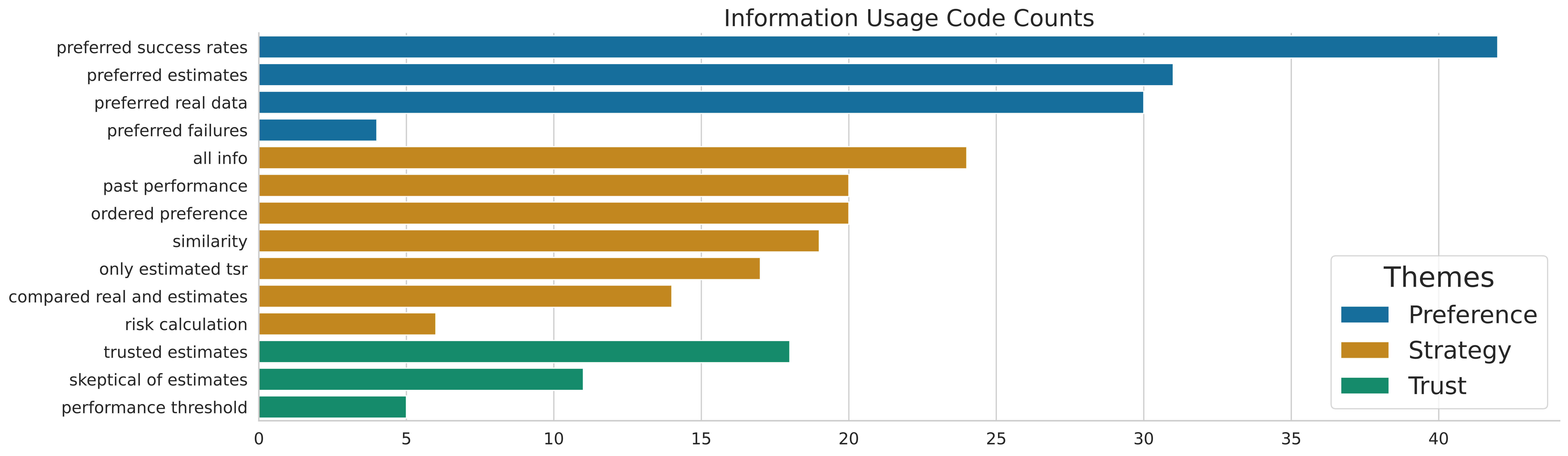}
\vspace{-4mm}
\caption{Information usage code counts from online study.}
\end{figure*}

\begin{figure*}[h]
\centering
\includegraphics[width=.99\textwidth]{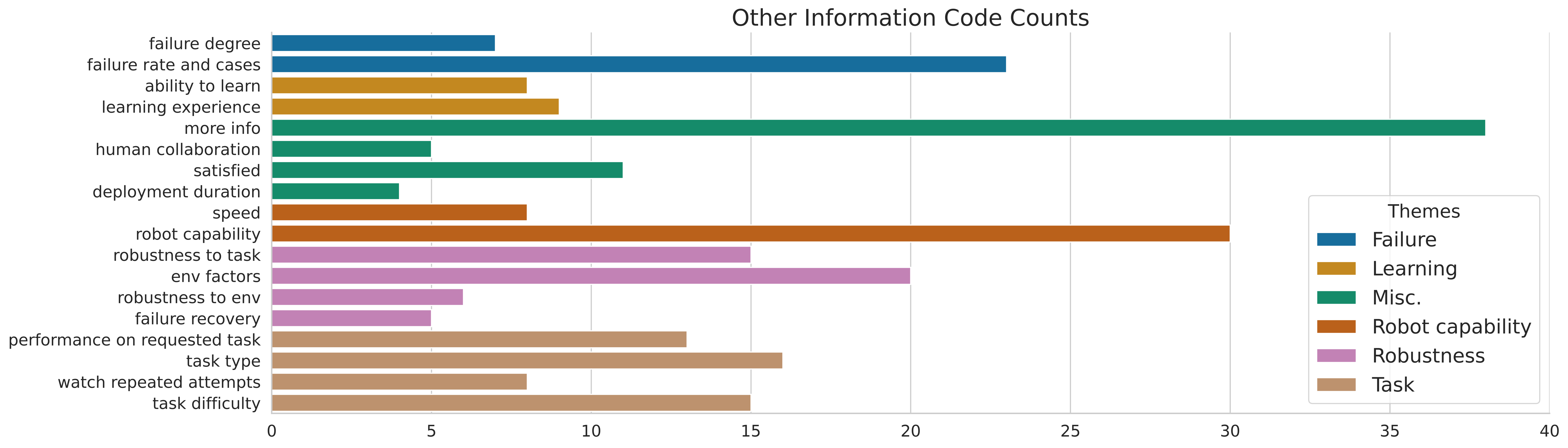}
\vspace{-4mm}
\caption{Other information code counts from online study.}
\end{figure*}

\begin{figure*}[h]
\centering
\includegraphics[width=.99\textwidth]{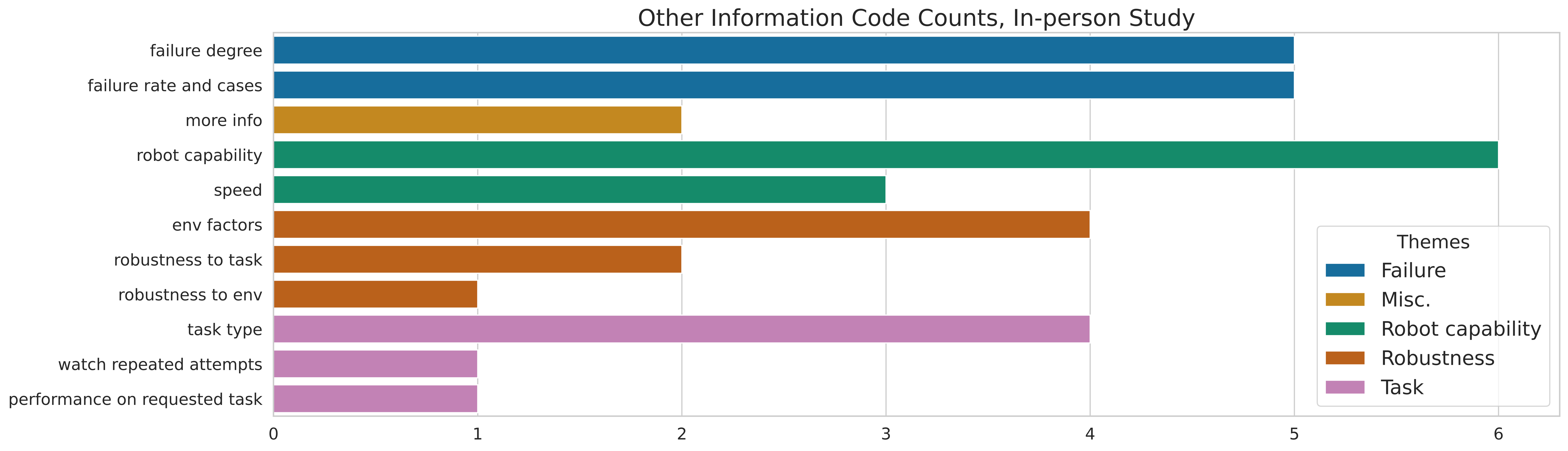}
\vspace{-4mm}
\caption{Other information code counts from in-person study.}
\end{figure*}

\newpage
\section{Common Demographic Information (Online Study)}

\begin{center}

\centering
\begin{tabular}{|l|l|l|}

\hline
Age &
  Gender &
  Robot Experience \\ \hline
\begin{tabular}[c]{@{}l@{}}\textit{18-24}: 16, \textit{25-34}: 40, \\ \textit{35-44}: 24, \textit{45-54}: 13,\\ \textit{55-64}: 7, \textit{65+}: 1\end{tabular} &
  \begin{tabular}[c]{@{}l@{}}Female: 54,\\ Male: 46,\\ Non-binary: 1\end{tabular} &
  \begin{tabular}[c]{@{}l@{}}None: 36, Slight: 38,\\ Moderate: 21, \\ Significant: 6\end{tabular} \\ \hline
\end{tabular}%

\label{table:demo}
\end{center}

\section{List of Questions Used}
Here is a list all of the questions asked to participants in the online and offline studies (excluding consent and background ). Questions labeled ``Likert'' indicate a 5-point Likert question ranging from ``strongly disagree'' to ``strongly agree.'' 

\subsection{Online study pre-task execution questions}
These are the questions asked to participants before seeing the robot attempt a task. Participants could see the description of the task, the task image, and the performance information presented in an information box. 
\begin{itemize}
    \item What is the reported robot's estimate of its success rate for this task? (manipulation check) \\
    \textit{response options:} 0-25\%, 26-50\%, 51-75\%, 76-100\%, Information not available

    \item I am confident the robot will be able to successfully perform this task. (Likert)

    \item I will be surprised if the robot fails. (Likert)

    \item I would feel comfortable with this robot performing this task in my own home. (Likert)

    \item I trust the robot to do the task on its own. (Likert)

    \item I would spend time improving the robot before letting it attempt this task. (Likert)

    \item I have enough information to assess the robot's ability to perform this task. (Likert)
\end{itemize}

\subsection{Online study post-task execution questions}
These are the questions asked to participants after seeing the robot attempt a task. Participants could still see the description of the task and performance information presented in an information box. They could also freely rewind and re-watch the task execution video.
\begin{itemize}
    \item The robot succeeded at the requested task. (Likert)

    \item I was surprised by the robot's behavior. (Likert)

    \item I received sufficient information to predict this outcome. (Likert)

    \item I would let the robot do the task again. (Likert)

    \item Do you have any other comments about the task, information, or robot behavior? (optional) (open-response)
\end{itemize}

\subsection{Online study post-study questions}
These are the questions asked after participants experienced all 16 robot tasks.
\begin{itemize}
    \item Please rank each type of information based on how useful it was. (drop-down menu ranking)
    
    \item Estimated task success rate for a requested task is useful for determining whether or not I want to use the robot. (Likert)

    \item Estimated failure cases for a requested task are useful for determining whether or not I want to use the robot. (Likert)

    \item Real task success rate for a similar task is useful for determining whether or not I want to use the robot. (Likert)

    \item Real task failure examples for a similar task are useful for determining whether or not I want to use the robot. (Likert)

    \item How did you use each type of information to make your decision? What made certain information more useful than other information? (open-response)

    \item What other types of information may be useful to you to decide when a robot can or cannot reliably perform a task? (open-response)
\end{itemize}

\subsection{In-Person study questions}
\begin{itemize}
    \item Please rank each type of information based on how useful it was. (drop-down menu ranking)
    
    \item Estimated task success rate for a requested task is useful for determining whether or not I want to use the robot. (Likert)

    \item Estimated failure cases for a requested task are useful for determining whether or not I want to use the robot. (Likert)

    \item Real task success rate for a similar task is useful for determining whether or not I want to use the robot. (Likert)

    \item Real task failure examples for a similar task are useful for determining whether or not I want to use the robot. (Likert)
    
    \item What other types of information may be useful to you to decide when a robot can or cannot reliably perform a task? (open-response)

    \item How would you decide whether or not the robot was good enough at a task to want one? (open-response)
    
\end{itemize}

\newpage

\section{Online study instructions and GUI}
Here we share images of the web page format shown to participants throughout the study. This includes the instruction page, pages for both a pre- and post- task execution, and the page for the post-study questionnaire.

\subsection{Study Instruction Page}
\begin{figure*}[h]
\centering
\includegraphics[width=.8\textwidth]{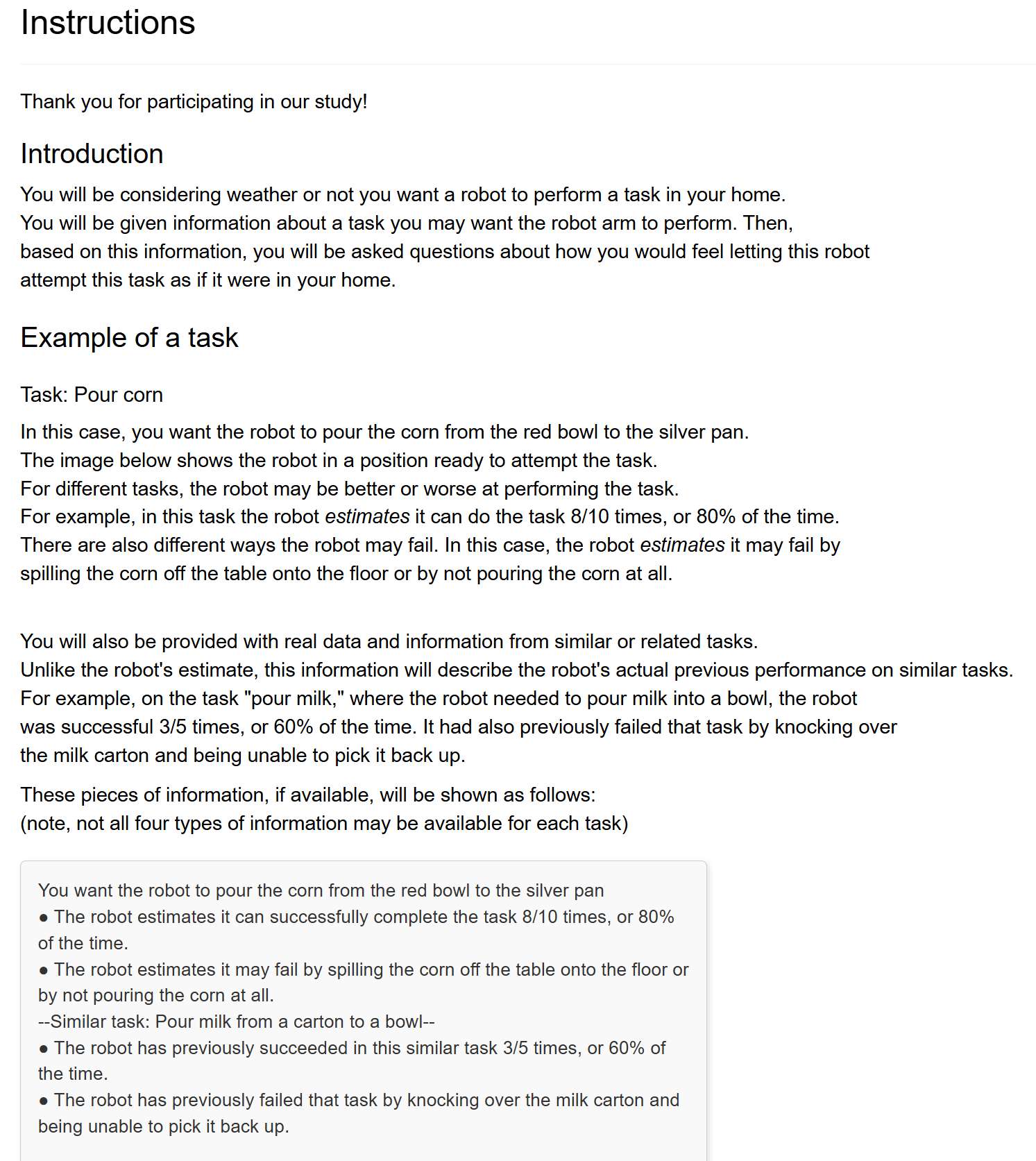}
\vspace{-4mm}
\caption{Instruction page (1/3)}
\end{figure*}
\newpage
\begin{figure*}[!htb]
\centering
\includegraphics[width=.8\textwidth]{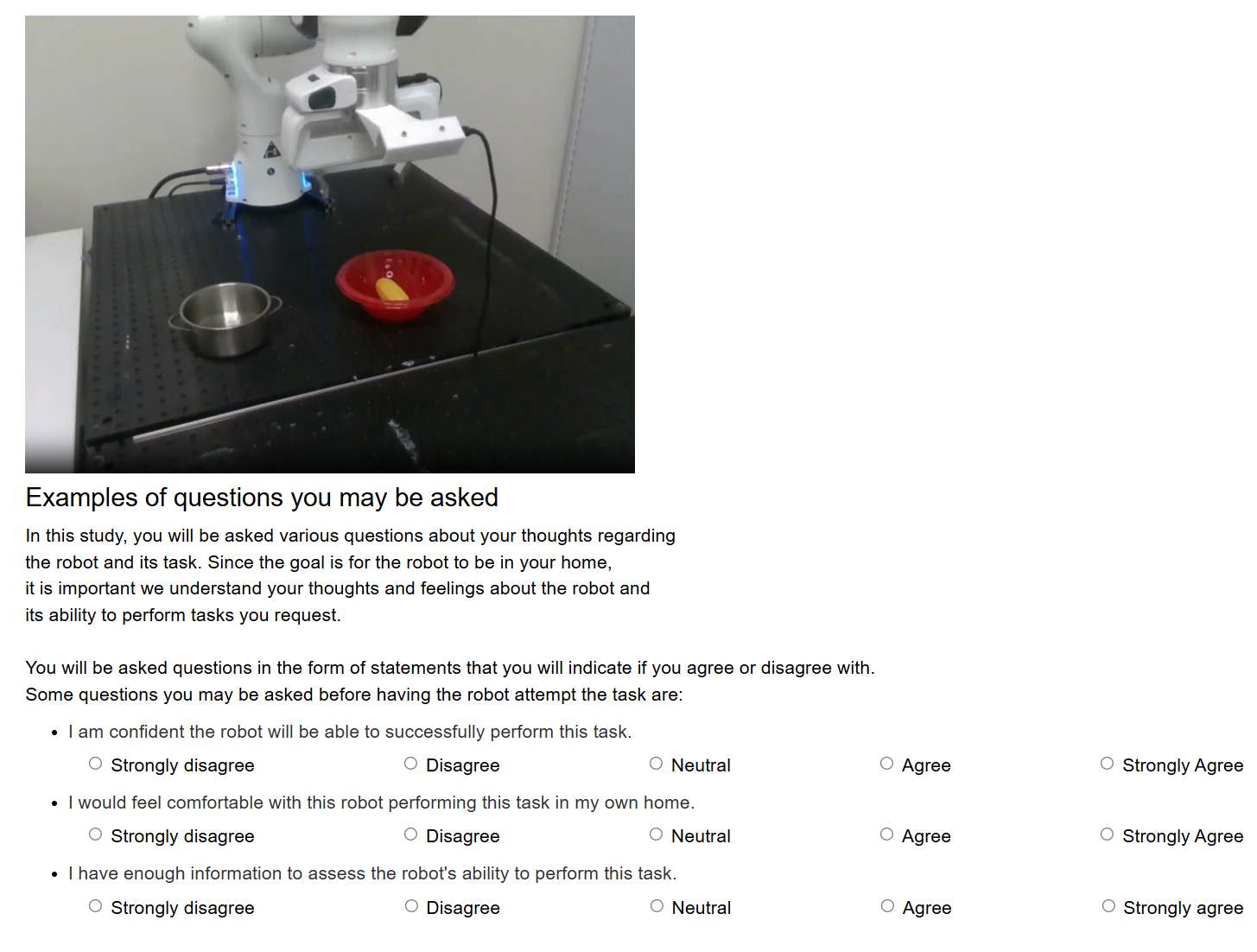}
\vspace{-4mm}
\caption{Instruction page (2/3)}
\end{figure*}
\newpage
\begin{figure*}[!h]
\centering
\includegraphics[width=.8\textwidth]{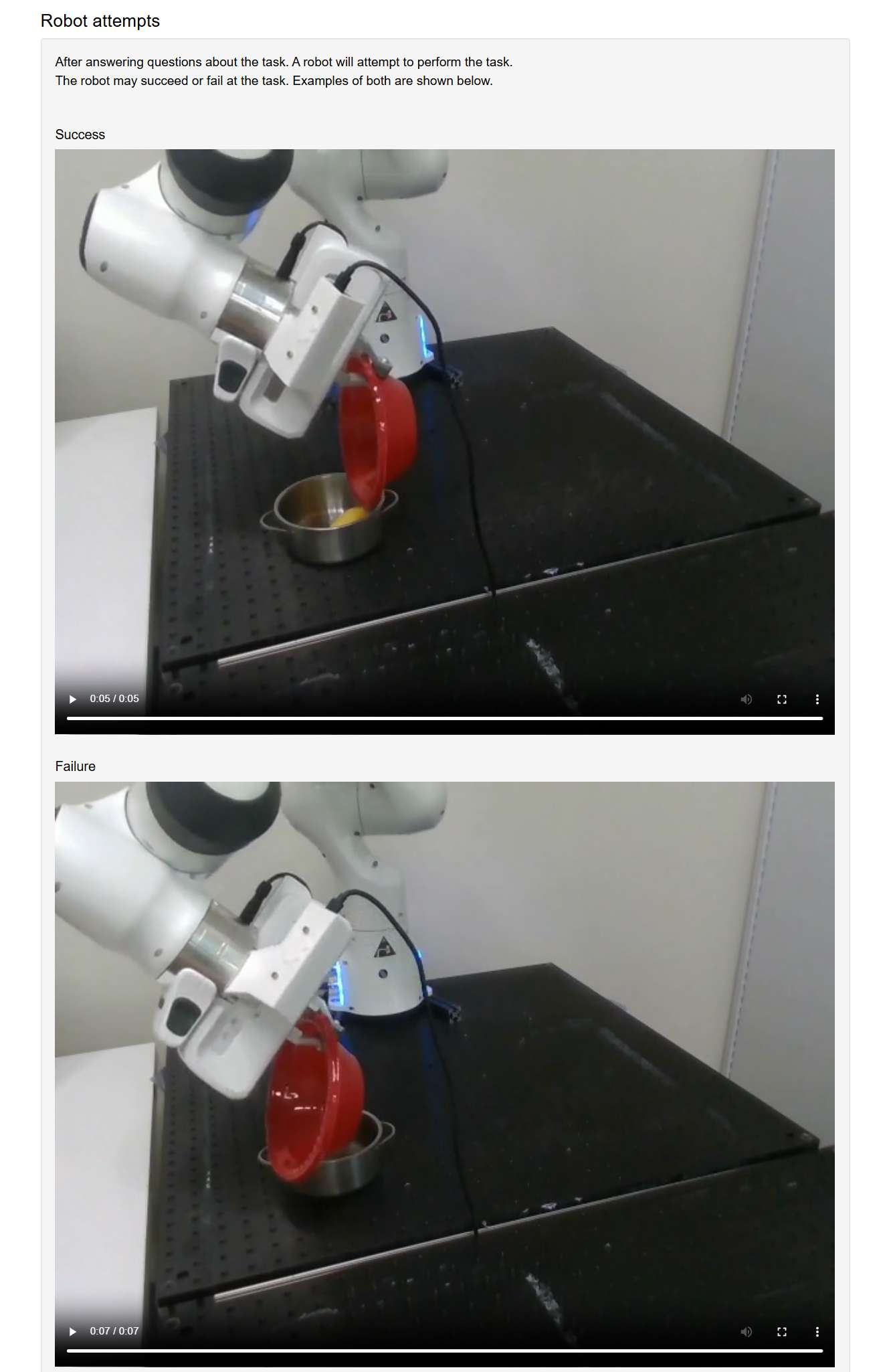}
\vspace{-4mm}
\caption{Instruction page (3/3)}
\end{figure*}

\newpage
\subsection{Task Pages}
\begin{figure*}[h]
\centering
\includegraphics[width=.85\textwidth]{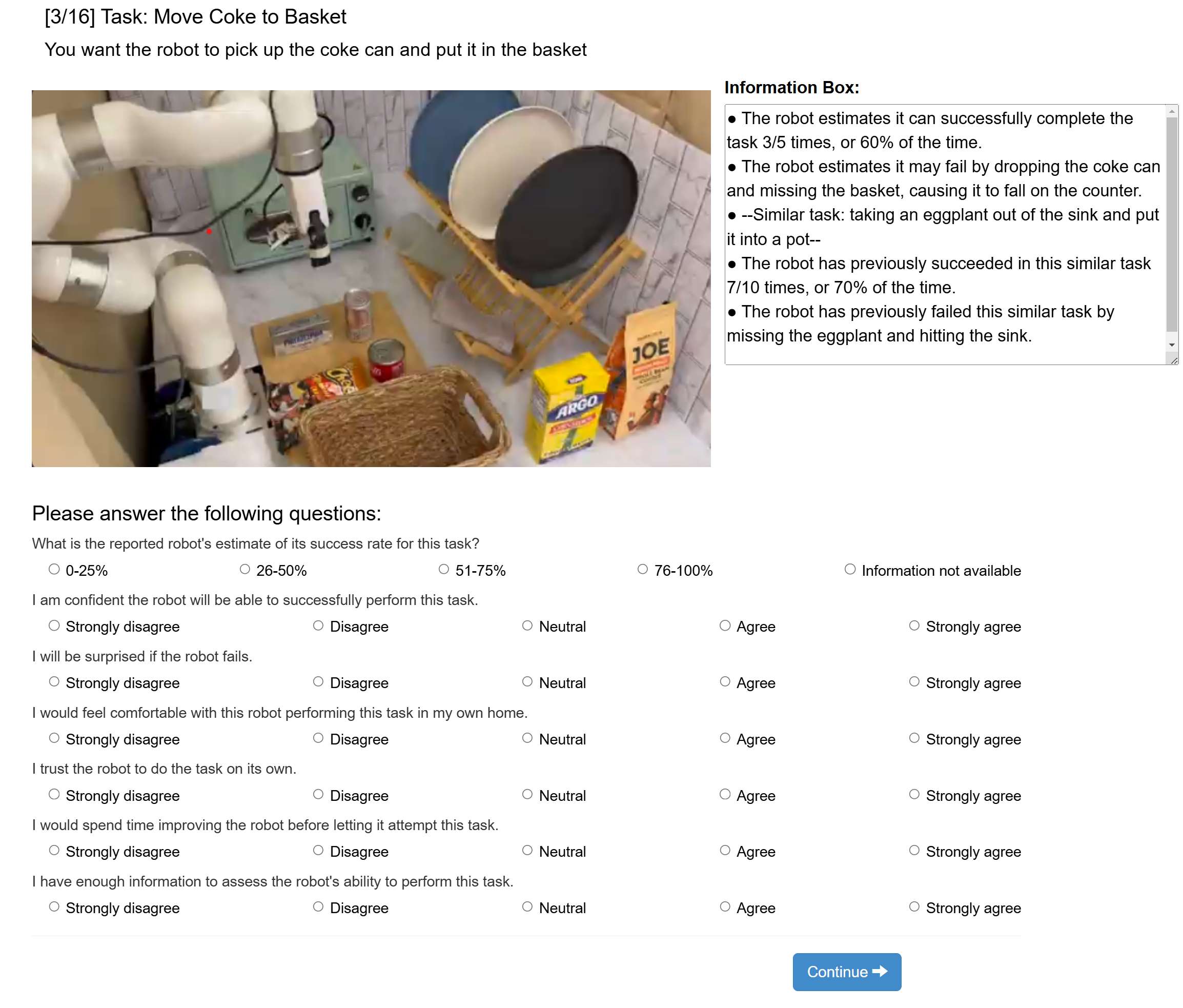}
\vspace{-4mm}
\caption{Pre-task execution questionnaire.}
\end{figure*}

\newpage
\begin{figure*}[h]
\centering
\includegraphics[width=.85\textwidth]{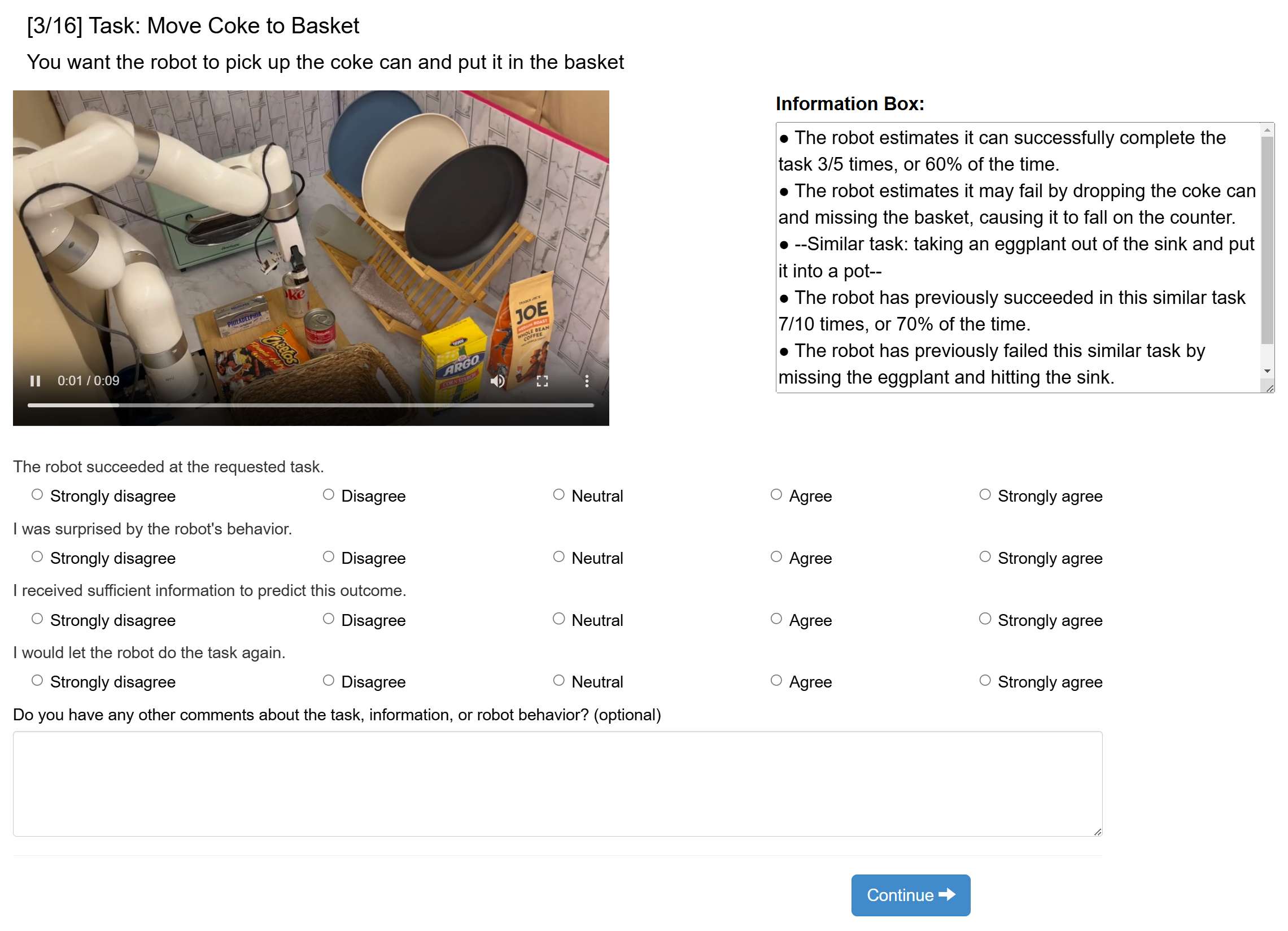}
\vspace{-4mm}
\caption{Post-task execution questionnaire.}
\end{figure*}

\newpage

\subsection{Post-study questionnaire}
\begin{figure*}[h]
\centering
\includegraphics[width=.85\textwidth]{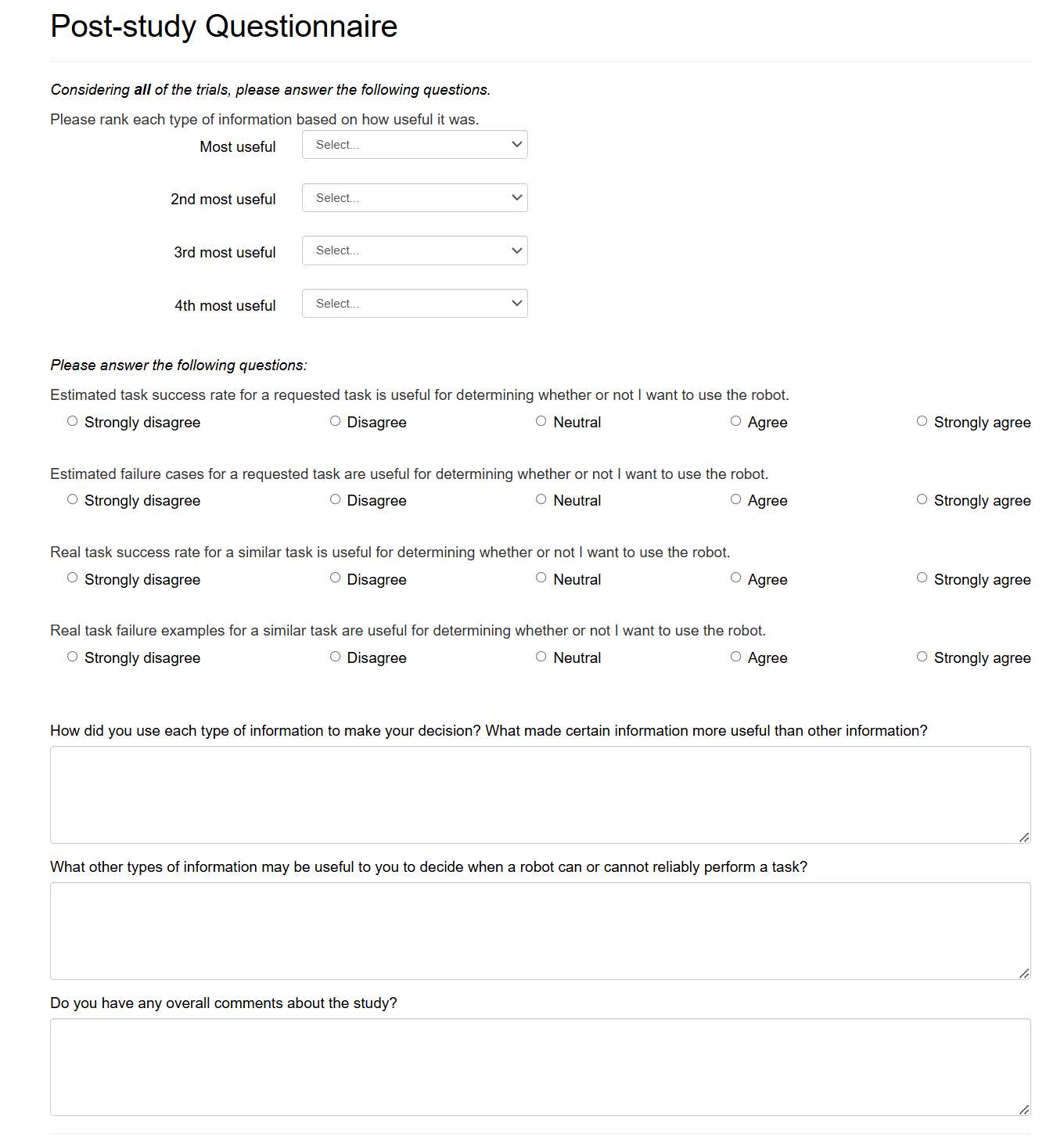}
\vspace{-4mm}
\caption{Post-study questionnaire.}
\end{figure*}

\end{document}